\documentclass[journal]{IEEEtran}

\usepackage[utf8]{inputenc} 
\usepackage[T1]{fontenc}    
\usepackage{hyperref}       
\usepackage{url}            
\usepackage{booktabs}       
\usepackage{amsfonts}       
\usepackage{amsmath}
\usepackage{empheq}
\usepackage{graphicx}
\usepackage{nicefrac}       
\usepackage{microtype}      
\usepackage[table]{xcolor}         
\usepackage{subfiles} 
\usepackage{bm}
\usepackage{stmaryrd}
\usepackage{enumitem}
\usepackage{wrapfig}
\usepackage{comment}
\usepackage{nicefrac}
\usepackage{balance}
\usepackage{comment}
\usepackage{tablefootnote}
\usepackage[linesnumbered,lined,ruled]{algorithm2e}
\usepackage{pifont}
\newcommand{\cmark}{\ding{51}}%

\SetCommentSty{mycommfont}

\title{Contact Models in Robotics:\\ a Comparative Analysis}

\definecolor{paleaqua}{rgb}{0.74, 0.83, 0.9}
\definecolor{pastelblue}{rgb}{0.68, 0.78, 0.81}

\DeclareMathOperator*{\argmin}{arg\,min}

\author{%
  Quentin Le Lidec\textsuperscript{1,\textdagger}, \and Wilson Jallet\textsuperscript{1,2}, \and Louis Montaut\textsuperscript{1,3}, \and Ivan Laptev\textsuperscript{1}, \and Cordelia Schmid\textsuperscript{1}, and \and Justin Carpentier\textsuperscript{1}%
  \thanks{%
        \textsuperscript{1}Inria - Département d'Informatique de l'\'Ecole normale supérieure, PSL Research University. Email: \texttt{firstname.lastname@inria.fr}}
    \thanks{%
        \textsuperscript{2}LAAS-CNRS, 7 av.~du Colonel Roche, 31400 Toulouse}
    \thanks{%
        \textsuperscript{3}Czech Institute of Informatics, Robotics and Cybernetics, Czech Technical University, Prague, Czech Republic}
    \thanks{%
        \textsuperscript{\textdagger}Corresponding author}
}

\begin{document}

\maketitle

\begin{abstract}
  Physics simulation is ubiquitous in robotics.
  Whether in model-based approaches (\textit{e.g.}, trajectory optimization), or model-free algorithms (\textit{e.g.}, reinforcement learning), physics simulators are a central component of modern control pipelines in robotics.
  Over the past decades, several robotic simulators have been developed, each with dedicated contact modeling assumptions and algorithmic solutions.
  In this article, we survey the main contact models and the associated numerical methods commonly used in robotics for simulating advanced robot motions involving contact interactions.
  In particular, we recall the physical laws underlying contacts and friction (\textit{i.e.}, Signorini condition, Coulomb's law, and the maximum dissipation principle), and how they are transcribed in current simulators.
  For each physics engine, we expose their inherent physical relaxations along with their limitations due to the numerical techniques employed.
  Based on our study, we propose theoretically grounded quantitative criteria on which we build benchmarks assessing both the physical and computational aspects of simulation.
  We support our work with an open-source and efficient C++ implementation of the existing algorithmic variations.
  Our results demonstrate that some approximations or algorithms commonly used in robotics can severely widen the reality gap and impact target applications.
  We hope this work will help motivate the development of new contact models, contact solvers, and robotic simulators in general, at the root of recent progress in motion generation in robotics.
\end{abstract}


\section{Introduction}
\begin{figure}
    \centering
    \includegraphics[width=.95\linewidth]{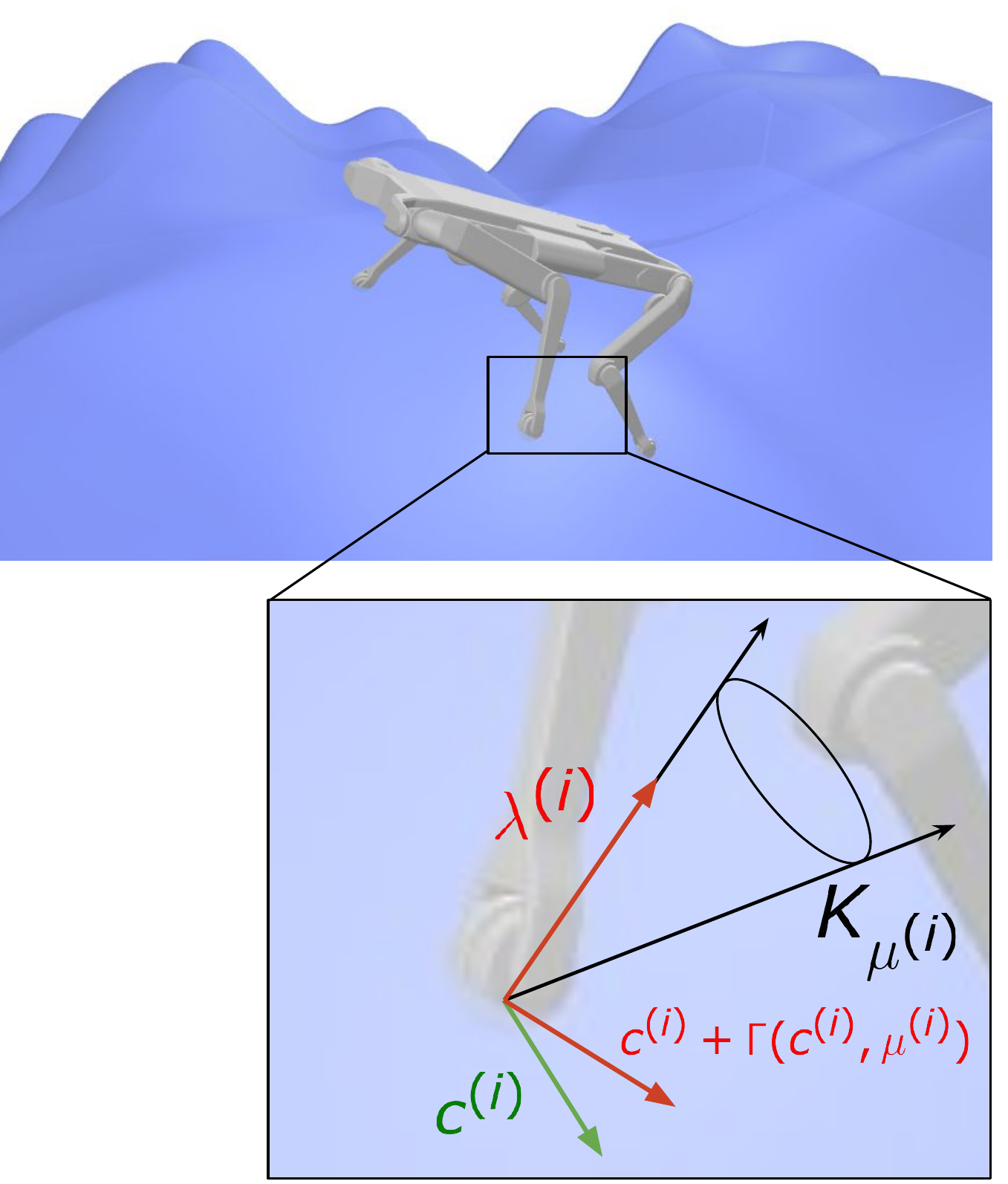}
    \caption{ \textbf{Illustration of the dynamics of frictional contacts} between rigid bodies, which are governed by the Signorini condition, Coulomb's law, and the maximum dissipation principle.
    Combining these three principles leads to the Non-linear Complementarity Problem \eqref{eq:NCP}.}
    \label{fig:solo_contact_scheme}
    \vspace{-0.5cm}
\end{figure}

\IEEEPARstart{S}{imulation} is a fundamental tool in robotics. 
Control algorithms, like trajectory optimization (TO) or model predictive control (MPC), rely on physics simulators to evaluate the dynamics of the controlled system.
Reinforcement Learning (RL) algorithms operate by trial and error and require a simulator to avoid time-consuming and costly failures on real hardware.
Robot co-design aims at finding optimal hardware design and morphology, and thus extensively relies on simulation to prevent tedious physical validation. 
In practice, roboticists also usually perform simulated safety checks before running a new controller on their robots.
These applications are evidence for a wide range of research areas in robotics where simulation is critical.

To be effective and valuable in practice, robot simulators must meet some fidelity or efficiency levels, depending on the use case.
For instance, trajectory optimization algorithms, \textit{e.g.}, iLQR\cite{li2004ilqr} or DDP \cite{ mayne1966second,tassa2012synthesis}, use physics simulation to evaluate the system dynamics and leverage finite differences or the recent advent of differentiable simulators \cite{carpentier2018analytical, de2018end,lelidec2021differentiable,werling2021fast, howelllecleach2022} to compute derivatives. 
If the solution lacks precision, the real and planned trajectories may quickly diverge, impacting \textit{de facto} the capacity of such control solutions to be deployed on real hardware.
To absorb such errors, the Model Predictive Control (MPC) \cite{diehl2006fast,mayne2014model} paradigm exploits state feedback by repeatedly running Optimal Control (OC) algorithms at high-frequency rates (e.g., 1kHz)~\cite{kleff2021high,dantec2022first}.
The frequency rate is one factor determining the robustness of this closed-loop algorithm to modeling errors and perturbations; thus, the efficiency of the simulation becomes critical. 
\IEEEpubidadjcol
Although RL~\cite{sutton2018reinforcement} is considered a model-free approach, physical models are still at work to generate the samples that are indispensable for learning control policies.
In fact, the vast number of required samples is the main bottleneck during training, as days or years of simulation, which corresponds to billions of calls to a simulator, are necessary \cite{tan2018sim,akkaya2019solving, hwangbo2019learning}.
Therefore, the efficiency of the simulator directly determines the computational and, thus, the energetic cost of learning a control policy.
Physical accuracy plays an important role after training as well, as more physically accurate simulations will result in a smaller reality gap to cross for the learned policy to transfer to a real robot \cite{tan2018sim}.

Many manipulation tasks can be tackled by assuming quasi-staticity and considering only a restricted variety of contact events \cite{mason1985robot,ponce1997computing}.
The recent robotics efforts, highlighted, for instance, by the athletic motions performed by the humanoid robots of Boston Dynamics~\cite{BD2022video}, focus on very dynamic tasks for which these simplification hypotheses cannot hold.
In fact, tasks like agile locomotion or dexterous manipulation require the robot to quickly plan and finely exploit, at best, the contact interactions with its environment to shape the movements \cite{mordatch2012discovery, posa2014direct, toussaint2018differentiable}.
In this respect, the ability to handle impacts and friction, physical phenomena at the core of contact interactions, becomes fundamental for robotic simulators.

Physics simulation is often considered a solved problem with several well-known simulators that are available off the shelf.
However, simulating a physical system raises several complex issues that are usually circumvented at the cost of approximations or costly computation.
When simulating a system evolving freely, rigid body dynamics algorithms~\cite{featherstone2014rigid,carpentier2019pinocchio} are now established as the way to go due to their high efficiency.
For robotics, one has to consider interactions through contact between the robot and its environment, thus constraining the movement.
However, due to the creation of the breaking of contacts along a trajectory, the dynamics switch from one mode to the other, making the problem of simulating a system with contacts and friction highly non-smooth~\cite{moreau1988mdp, jean1999non, acary2011contact}.
Numerical integration schemes for non-smooth systems fall into two main categories: event-driven and time-stepping methods \cite{brogliato2002numerical}.
Most modern robotics simulators are part of the latter category because predicting collisions is intractable due to the complexity of the scenes. 
Therefore, we will restrict our study to this type of method.

More precisely, contact dynamics between rigid objects are governed by three main principles: the Signorini condition specifies the unilaterality nature of contact interactions, while Coulomb's law of friction and the maximum dissipation principle (MDP) of Moreau state that friction force should lie inside a second-order cone and oppose the movement.
Altogether, these three principles correspond to a so-called nonlinear complementarity problem\,(NCP).
The complementarity constraints define a non-convex set while being non-smooth, this problem is difficult to solve in general~\cite{brogliato2002numerical}.

Historically, the Open Dynamic Engine (ODE)\,\cite{ode:2008} is one of the first open-source simulators with a large impact on the community, which was then followed by Bullet\,\cite{coumans2021}. 
Both of them, in their original version, relied on maximal coordinates to depict the state of the objects, and kinematic constraints imposed by the articulations are tackled explicitly. 
Such a choice leads to large-dimensional problems to solve, impacting \textit{de facto} the computational performances.
To lower the computational burden, alternative simulators rooted in generalized coordinates, like DART\,\cite{DART} and MuJoCo\,\cite{todorov2012mujoco}, appeared shortly after.
Since then, Bullet also made this choice the default one.
In practice, these simulators are rarely used to tackle engineering problems but rather as physics engines for graphical purposes (Bullet) or research in the RL community (MuJoCo).
More recently, RaiSim \cite{raisim} and Drake \cite{drake} were developed as robotic-driven software.
RaiSim\,\cite{raisim} emerged as one of the first simulators enabling RL policies to transfer to real quadrupedal robots. 
Its implementation being closed source, we provide what constitutes, to the best of our knowledge, the first in-depth study and open-source re-implementation of this contact solver.
Drake also demonstrated some promising results on challenging manipulation \cite{castro2022unconstrained} as regards the sim-to-real requirements.
Still today, the number of alternative algorithms available is growing fast, in an effort to improve the properties of the existing ones, in terms of accuracy and robustness~\cite{kaufman2008staggered, macklin2019non, enzenhofer2019efficient, lelidec2021differentiable, ferguson2021intersection, castro2022unconstrained, howelllecleach2022}.
In a parallel line of work, Isaac Gym~\cite{makoviychuk2021isaac} and Brax~\cite{freeman2021brax} simulators use elementary contact models and rather focus on exploiting the parallelization abilities from GPUs or TPUs for batch computation.

In general, these simulators differ at their very core: one should be aware of the contact modeling embedded in the simulator they are using and how it can impact the applications they aim at.
Some high-level benchmarks of simulators exist \cite{erez2015bench}, evaluating the whole simulation pipeline and its multiple internal routines, \textit{e.g.}, rigid-body dynamics algorithms, collision detection, and contact problem-solving.
Our work closely relates to \cite{Horak2019OnTS}. It separately assesses the various contact models and their associated algorithms. 
We achieve this by decoupling the contact models from their implementations and re-implemented the solvers with a unique back-end based on the Pinocchio toolbox~\cite{carpentier2019pinocchio,pinocchioweb} for evaluating the dynamic quantities and on HPP-FCL\,\cite{hppfclweb,mirabel2016hpp,montaut2022collacc} for computing the collisions.
We pursue the effort of \cite{Horak2019OnTS} by studying recent algorithms and adding advanced evaluation criteria.
Our experiments are done in both illustrative and realistic robotics setups.

\noindent We make the following contributions:
\begin{itemize}[label={$\hookrightarrow$}]
    \item we make a detailed survey of contact models and their associated algorithms, including established and more recent robotics simulators;
    \item we expose the main limitations of existing simulators by inspecting both the physical approximations and the numerical methods that are at work;
    \item we develop an open source and generic implementation of the main robotic contact solvers in C++;
    \item based on our implementation and the theoretical study, we propose quantitative criteria that allow performing an in-depth evaluation of both physical and computational aspects of contact models and solvers.
    \item we explore the impacts of the simulation choices on the practical application of MPC for quadruped locomotion.
\end{itemize}

The article is organized as follows: we first recall the background of contact simulation: the physical principles behind contact modeling (Sec.~\ref{sec:models}) and the numerical algorithms allowing us to solve the resulting equations (Sec.~\ref{sec:solvers}).
In the experimental part (Sec.~\ref{sec:exp}), we propose an exhaustive empirical evaluation of the various existing contact models and solvers to assess both their physicality (Sec.~\ref{sec:exp_phys}), self-consistency (Sec.~\ref{sec:exp_cons}) and computational efficiency (Sec.~\ref{sec:exp_perf}).
At last, Sec.~\ref{sec:exp_mpc} investigates the consequences of the contact models in the context of quadruped locomotion.
It is finally worth mentioning that the authors are linked to the Pinocchio and HPP-FCL open-source projects.
\section{Rigid contact modelling} \label{sec:models}

We start by stating the physical principles commonly admitted for rigid body simulation with point contact.
If these principles remain hypothetical and can still be discussed, they have been, in general, empirically tested and are arguably better than their relaxations.
Once the modeling is done, we transcribe these physical laws into a numerical problem, which should be solved via optimization-based techniques to simulate a system with contacts and frictions.
We also present the various open-source tools that allow computing all the intermediate quantities necessary to build a physics simulator.

In this paper, we describe the state of a system with its generalized coordinates \mbox{$q\in \mathcal{Q} \cong \mathbb{R}^{n_q}$}.
We denote by \mbox{$v  \in \mathcal{T}_q\mathcal{Q} = \mathbb{R}^{n_v}$} the joint velocity, where $  \mathcal{T}_q\mathcal{Q}$ is the tangent space of $\mathcal{Q}$.\\

\noindent
\textbf{Free motion.} The principle of least constraint \cite{udwadia1992new,bruyninckx2000gauss,redon:hal-01147672} induces the celebrated equations of motion:
\begin{equation}
    \label{eq:newton}
    M(q)\dot{v} +b(q,v) = \tau
\end{equation}
where $M\in{R}^{n_v\times n_v}$ represents the joint space inertia matrix of the system, $b(q,v)$ accounts for the centrifugal and Coriolis effects, and for the generalized gravity.
This Lagrangian equation of motion naturally accounts for the kinematic constraints induced by the articulations of the rigid-body dynamical system.
When applied to a robot, \textit{i.e.}, a system of multiple rigid bodies, the inertia matrix $M$ becomes sparse.
Rigid body dynamic algorithms exploit this sparsity at best \cite{featherstone2014rigid,carpentier2019pinocchio} making it possible to compute the free acceleration in a few microseconds on modern CPUs for robots as complex as a 36-dof humanoid.
As done by time-stepping approaches \cite{brogliato2002numerical}, we will express the problem in terms of velocities rather than accelerations, thus discretizing \eqref{eq:newton} into:
\begin{equation}
    \label{eq:discrete_newton}
    M(q^t) v^{t+1} = M(q^t) v^t + (\tau^t - b(q^t,v^t)) \Delta t
\end{equation}
which corresponds to a semi-implicit Euler integration scheme \cite{todorov2011convex}.
More advanced implicit integrators \cite{todorov2012mujoco,castro2022unconstrained,howelllecleach2022} come with stability guarantees even in the presence of stiff forces.
However, as time-stepping schemes, their order of integration is inherently degraded due to the non-smoothness of the dynamics \cite{studer2008step}.
For this reason, we restrict our study to a simple scheme, as integrators are not the main focus of this work.
In the following, we often drop the instant at which quantities are evaluated for readability purposes.
We denote the free velocity $v^f$,  which is defined as the solution of \eqref{eq:discrete_newton}.\\

\noindent
\textbf{Bilateral contact.} When the system is subject to constraints, \textit{e.g.} kinematic loop closures or anchor points, it is convenient to represent them implicitly:
\begin{equation}
    \Phi(q) = 0. \label{eq:bilateral_constraint}
\end{equation}
 where $\Phi: \mathbb{R}^{n_q} \mapsto \mathbb{R}^m$ is a holonomic constraint function of dimension $m$, which depends on the nature of the constraint.
For solving, it is more practical to proceed to an index reduction \cite{Campbell1995TheIO} by differentiating \eqref{eq:bilateral_constraint} w.r.t. time, in order to express it as a constraint on joint velocities:
\begin{equation}
    c  - c^*= 0. \label{eq:bilateral_constraint_vel}
\end{equation}
where $c = J(q^t)v^{t+1} \in \mathbb{R}^m$ is the constraint velocity, $J = \nicefrac{\partial \Phi}{\partial q}$ is the constraint Jacobian explicitly formed at time $t$, which can be computed efficiently via rigid body dynamics algorithms~\cite{featherstone2014rigid,carpentier2021proximal}; and $c^*$ is the reference velocity which stabilizes the constraint.
Such a constraint \eqref{eq:bilateral_constraint_vel} is enforced by the action of the environment on the system via the contact vector impulse $\lambda \in \mathbb{R}^{m}$.
These considerations lead to Gauss's principle of least constraint\,\cite{gauss1829neues,udwadia1992new}. 
By duality, the contact impulses are spanned by the transpose of the constraint Jacobian and should be incorporated in the Lagrangian equations \eqref{eq:discrete_newton} via:
\begin{equation}
    \label{eq:contact_newton}
    M v^{t+1} = Mv_f + J^\top \lambda
\end{equation}
Regarding bilateral contacts, the contact efforts, corresponding to the Lagrange multipliers associated with the constraint \eqref{eq:bilateral_constraint}, are unconstrained.
If a bilateral constraint is well suited to model kinematic closures, it is not to model interactions between the robot and its environment, which are better represented by unilateral contacts.
This paper focuses on the latter, for which we provide a more detailed presentation.\\

\begin{figure}
    \centering
    \includegraphics[width=.45\linewidth]{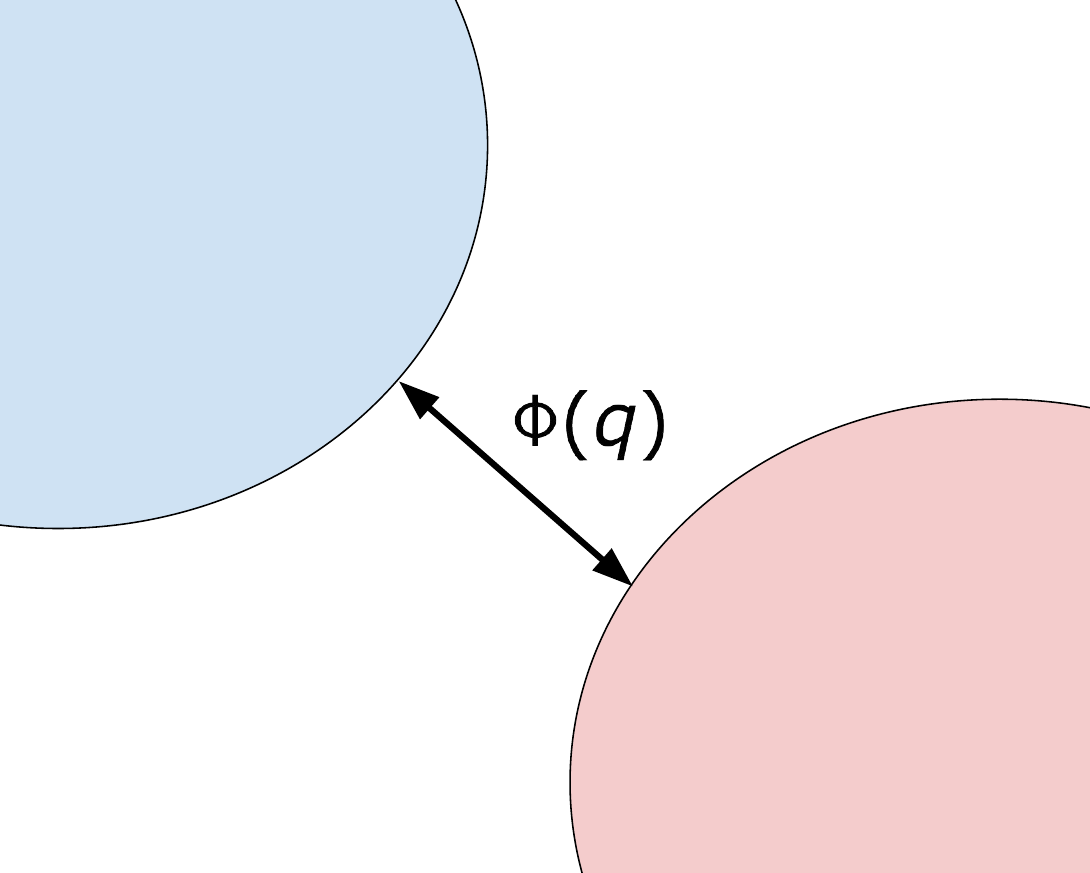}
    \includegraphics[width=.45\linewidth]{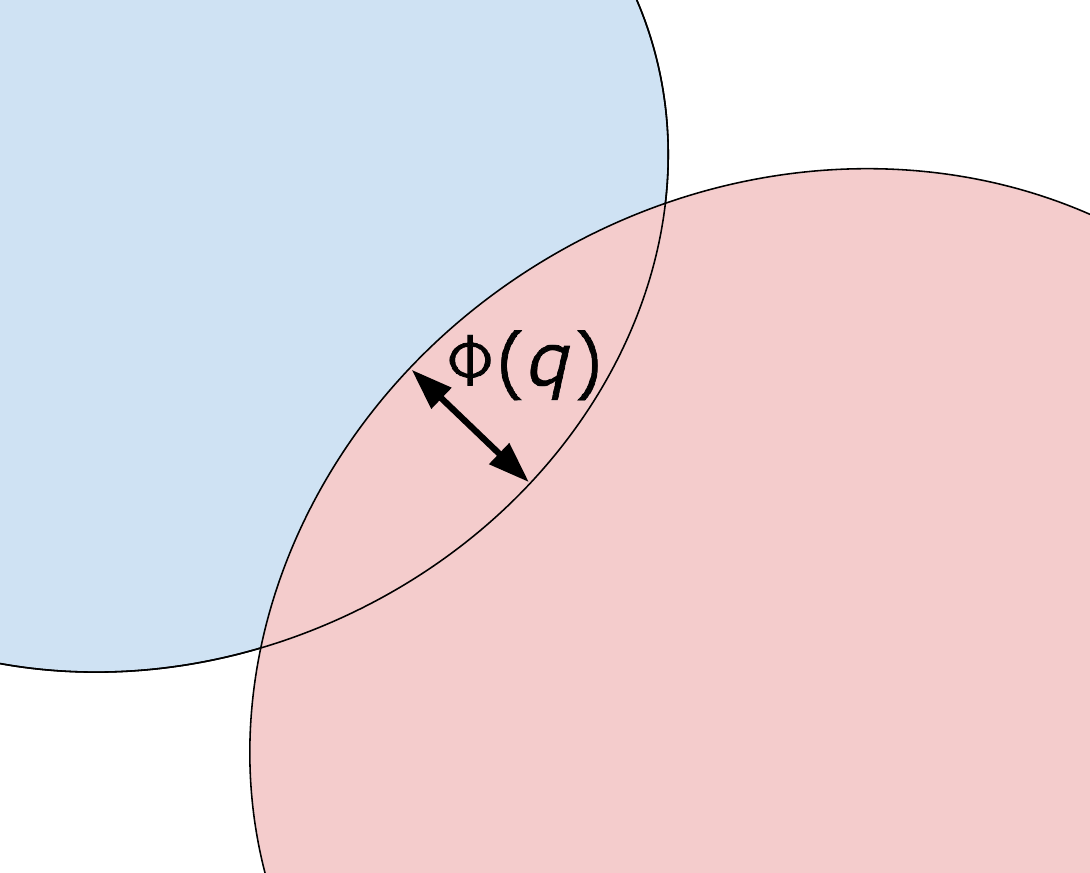}
    \caption{ \textbf{The separation vector $\Phi$} allows formulating the non-penetration constraint, which leads to the \textit{Signorini condition} \eqref{eq:signorini_cond}. This vector is computed by the GJK or EPA algorithms, which are internal blocks of the simulator. We refer to~\cite{montaut2022collision} for a tutorial introduction on the topic.}
    \label{fig:phi_function}
    \vspace{-0.0cm}
\end{figure}

\begin{figure}
    \centering
    \includegraphics[width=.45\linewidth]{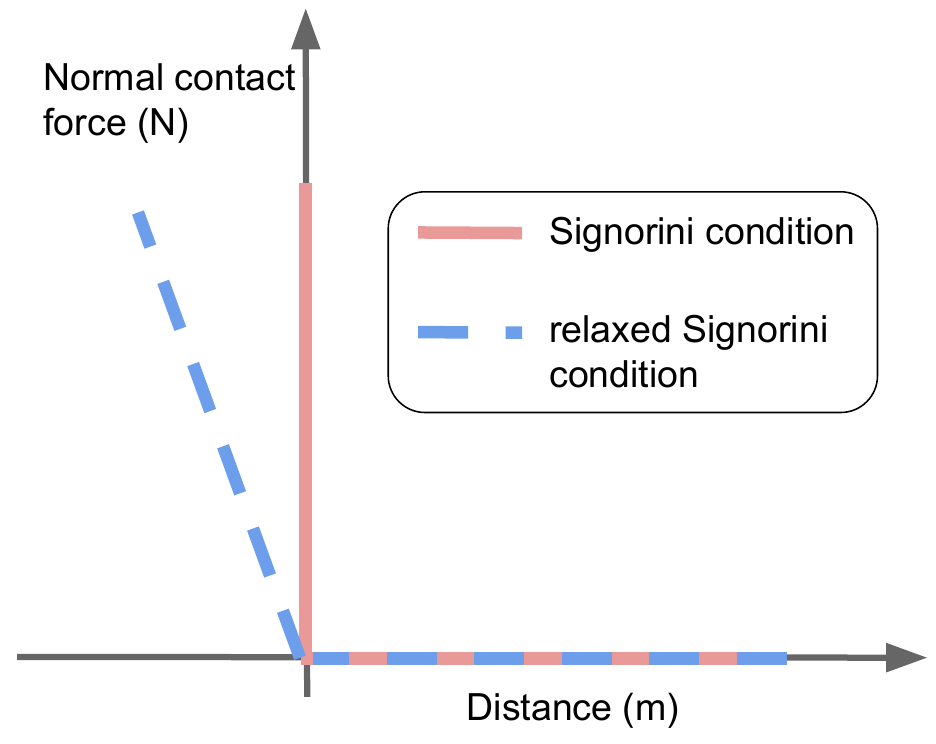}
    \includegraphics[width=.45\linewidth]{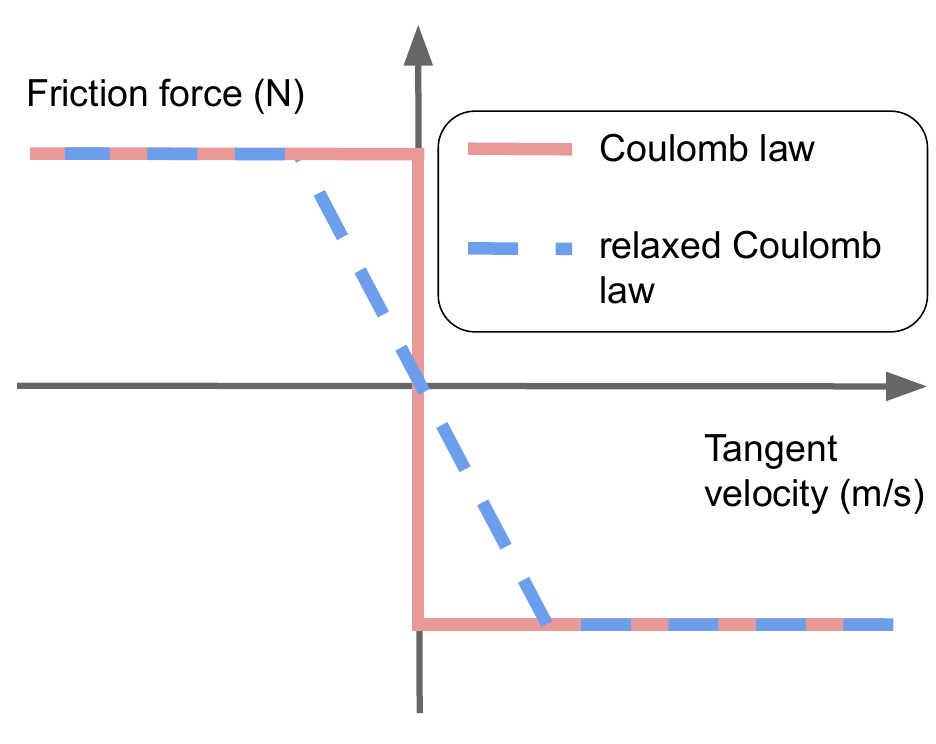}
    \caption{Both the \textit{Signorini condition} (\textbf{Left}) and Coulomb's law (\textbf{Right}) induce infinitely steep graphs, which make the contact problem hard to solve.}
    \label{fig:sig_coulomb}
    \vspace{-0.0cm}
\end{figure}

\noindent
\textbf{Unilateral contact.} When a system is in contact with its environment, the non-penetration constraint enforces the signed distance between the two objects to be non-negative \cite{pfeiffer2000multibody}.
Defining the separation vector as the vector of minimum norm separating two shapes in contact \cite{ericson2004real,montaut2022collision}(Fig.~\ref{fig:phi_function}), the signed distance function corresponds to its normal component.
By overloading the notation of the bilateral case, the constraint function $\Phi$ now maps to the separation vector (Fig.~\ref{fig:phi_function}) and describes a unilateral constraint:
\begin{equation}
    \Phi (q)_N \geq 0 \label{eq:non_penetration}
\end{equation}
where $\Phi(q)\in \mathbb{R}^{3n_c}$, $n_c$ is the number of contacts; the subscripts $N$ and $T$ respectively account for the normal and tangential components. 
In practice, $\Phi$ can be computed efficiently via the Gilbert-Johnson-Keerthi (GJK)\,\cite{gilbert1988fast,montaut2022collacc} and the Expanding Polytope Algorithm (EPA) algorithms\cite{ericsonRealTimeCollisionDetection}. 
GJK operates on convex shapes, but non-convex shapes can also be handled by proceeding to decomposition into convex sub-shapes \cite{mamou2009simple} during an offline preprocessing step.
To ease the solving, one can write\,\eqref{eq:non_penetration} in terms of velocities, and supposing that shapes are in contact, \textit{i.e.} $\Phi (q^t)_N \leq 0$, the Taylor expansion of the condition \eqref{eq:non_penetration} leads to:
\begin{equation}
    \label{eq:no_pen}
    c_N  - c^*_N\geq 0 
\end{equation}
where $c = J(q^t)v^{t+1} \in \mathbb{R}^{3n_c}$ is the velocity of contact points. 
It should be noted that J is evaluated at $q^t$ as it avoids computing $\Phi$ and its Jacobian several times when solving for $q^{t+1}$ and $v^{t+1}$ which significantly decreases the computational burden.
We explain later how $c^*_N$ is set to model physical effects or improve the numerical accuracy of the solutions.
As in the bilateral case, the transpose of the contact Jacobian $J$ spans the contact forces, which leads again to \eqref{eq:contact_newton} the constrained equations of motion.
Unlike the bilateral case, unilateral contacts constrain the possible contact impulses $\lambda$.
In a frictionless situation, the tangential forces are null, which implies that $\lambda_T=0$.
In addition, the contact forces $\lambda$ can only be repulsive \textit{i.e.}, they should not act in a glue-like fashion (the environment can only push and not pull on the feet of a legged robot) and, thus, are forced to be non-negative.
An impulse cannot occur when an object takes off, \textit{i.e.}, the normal velocity and impulse cannot be non-null simultaneously.
Combining these conditions, we obtain the so called \textit{Signorini condition} \cite{signorini1959questioni} at the velocity level \cite{moreau1988mdp}: 
\begin{equation}
    \label{eq:signorini_cond}
     0\leq \lambda_N \perp c_N - c_N^* \geq 0 .
\end{equation}
where $a \perp b$ for vectors $a$ and $b$ means $a^\top b = 0$.
However, such a condition does not define a mapping between $\lambda_N$ and $c_N$, \textit{i.e.}, the contact forces are not a function of the penetration error.
Indeed, their representation is an infinitely steep graph that may be relaxed into a mapping via a spring damper accounting for local deformation of the materials (see Fig.~\ref{fig:sig_coulomb}).
Substituting, $v^{t+1}$ by its expression from the Lagrangian equations \eqref{eq:contact_newton}, we obtain a Linear Complementarity Problem (LCP) \cite{cottle2009lcp}:
\begin{equation}
    \label{eq:LCP}
     0\leq \lambda_N \perp (G \lambda + g)_N - c_N^*\geq 0
\end{equation}
where $G = J M^{-1} J^\top$ is the so-called Delassus matrix, and $g=Jv^f$ is the free velocity of contact points (the velocity of the contact points in the unconstrained cases).
It is worth mentioning at this stage that several approaches~\cite{featherstone2014rigid,wensing2012reduced,carpentier2021proximal} have been developed in the computational dynamics and robotics literature to efficiently evaluate the Delassus matrix.

In the case of rigid bodies, the reference velocity $c_N^*$ can be set to $ \frac{\Phi(q^t)}{\Delta t}$ to complete the Taylor expansion of \eqref{eq:non_penetration}.
However, adding bias terms to this velocity may be useful to improve modeling on both physical and numerical aspects.
A first benefit is the possibility of accounting for impacts that may occur when two objects collide with non-null relative normal velocity.
The most common impact law stipulates to introduce a bias term $- e c^t$ where $e$ is the restitution coefficient, which adjusts the quantity of energy dissipated during the collision.
When time-stepping methods are employed, one cannot avoid penetration errors, \textit{i.e.} $\Phi(q)_N < 0$, without using stabilization by reprojection techniques \cite{bauchau2008review} which are computationally expensive to use in robotics due to the cost of detecting a collision.
However, it is still possible to prevent these errors from dramatically growing over time via a Baumgarte correction~\cite{baumgarte1972stabilization} which adds $k_B \max(0,-\Phi(q^t)_N)$ to the reference velocity $c^*$ and where the Baumgarte coefficient $k_B$ is set to be proportional to $ \frac{1}{\Delta t}$.

In addition, in many cases in robotics, Delassus' matrix $G$ is rank deficient.
Such physical systems are said to be hyperstatic, and because $\mathrm{rank}(J) > n_v$, several $\lambda$ values may lead to the same trajectory.
This under-determination can be circumvented by relaxing the rigid-body hypothesis, \textit{e.g.} the \textit{Signorini condition}, and considering compliant contacts via a reference velocity linearly depending on $\lambda$ as represented in Fig.~\ref{fig:sig_coulomb}.
Indeed, by adding $- R \lambda$ to $c^*$ where $R$ is a diagonal matrix with non-null and positive elements only on the normal components, called compliance and whose value is a property of the material, the original Delassus matrix $G$ is replaced by the damped matrix $\Tilde{G} = G + R$ which is full rank.
At this stage, one should note that the physical compliance acts on the conditioning in an equivalent way to a numerical regularization.\\

\noindent
\textbf{Friction phenomena} are at the core of contact modeling, as they precisely enable manipulation or locomotion tasks.
Coulomb's law for dry friction represents the most common way to model friction forces.
This phenomenological law states that the maximum friction forces $\| \lambda_T \|$ should be proportional to the normal contact forces $\lambda_N$ and the friction coefficient $\mu$.
Mathematically, this suggests that contact forces should lie inside an ice cream cone whose aperture is set by the coefficient of friction $\mu$:
\begin{equation}
\label{eq:coulomb_law}
    \lambda \in K_\mu = \prod_{i=1}^{n_c} K_{\mu^{(i)}}
\end{equation}
where the product is Cartesian, the superscript $(i)$ refers to the i$^{th}$ contact point and $K_{\mu^{(i)}}  = \left \{ \lambda | \lambda \in \mathbb{R}^3, \ \lambda_N \geq 0, \|\lambda_T\|_2 \leq \mu^{(i)} \lambda_N \right \}$.
Additionally, when sliding occurs, the maximum dissipation principle formulated by Jean-Jacques Moreau~\cite{moreau1988mdp} implies that the frictional forces should maximize the dissipated power:
\begin{equation}
    \label{eq:socp_mdp}
    \forall i, \ \lambda_T^{(i)} = \argmin_{\|\gamma_T \| \leq \mu^{(i)} \lambda_N^{(i)}} \gamma_T ^\top c_T^{(i)}
\end{equation}
whose optimality conditions yield the following equation in the sliding case:
\begin{equation}
    \label{eq:mdp}
    \forall i, \ \lambda_T^{(i)} = - \mu^{(i)}\, \lambda_N^{(i)} \frac{c_T^{(i)}}{\|c_T^{(i)}\|}, \mathrm{if} \ \|c_T^{(i)}\|>0.
\end{equation}
As for the \textit{Signorini condition}, Coulomb's law does not describe a mapping but an infinitely steep graph (Fig.~\ref{fig:sig_coulomb}). 
Relaxing this law via viscous frictions, \textit{i.e.}, assuming the tangent contact forces to be proportional to the tangent velocities, allows defining a mapping between $\lambda_T$ and $c_T$.\\

\noindent
\textbf{The Non-linear Complementarity Problem.}
Combining the Coulomb's law for friction with the \textit{Signorini condition} evoked earlier, we finally get three distinct cases corresponding to a sticking contact point \eqref{eq:stiction}, a sliding contact point \eqref{eq:sliding} or a take-off \eqref{eq:take-off}:
\begin{subequations}
\label{eq:disjunctive_form}
    \begin{empheq}[left={\empheqlbrace\,}]{align}
      \label{eq:stiction}
        \lambda^{(i)} \in K_{\mu^{(i)}}, \mathrm{if}\ c^{(i)} = 0\\ 
        \label{eq:take-off}
        \lambda^{(i)} = 0, \mathrm{if}\  c_N^{(i)} > 0\\
        \label{eq:sliding}
        \lambda^{(i)} \in \partial K_{\mu^{(i)}}, \exists \alpha > 0, \lambda_T^{(i)} = - \alpha c_T^{(i)} \ \mathrm{otherwise}.
    \end{empheq}
\end{subequations}
where $\partial K$ indicates the boundary of the cone.
The equations \eqref{eq:disjunctive_form} are referred to as the disjunctive formulation of the contact problem.
However, such a formulation is unsuitable in practice for solving, as the switching condition depends on the contact point velocity $c$.
As this quantity is an unknown of the problem, one cannot know in which case of \eqref{eq:disjunctive_form} one is standing.
For this reason, the problem is often reformulated as a nonlinear complementarity problem (NCP).
Indeed, using de Saxc\'{e}'s bipotential function~\cite{desaxce1998bipotential} defined as:
\begin{equation}
    \Gamma : (c,\mu) \in \mathbb{R}^3 \times \mathbb{R} \mapsto \left[0,0, \mu \|c_T \|_2 \right]
\end{equation}
one can show that \eqref{eq:disjunctive_form} is equivalent to the following \cite{acary2011contact,acary2017contact} (Fig.~\ref{fig:solo_contact_scheme}):
\begin{equation}
\label{eq:NCP}
    \forall i, K_{\mu^{(i)}} \ni \lambda^{(i)} \perp c^{(i)} + \Gamma\left(c^{(i)}, \mu^{(i)}\right) \in K_{\mu^{(i)}}^*.
\end{equation}
In \eqref{eq:NCP},~$K^*_{\mu}$ refers to the dual cone of~$K_{\mu}$, such that if~$\lambda \in K_{\mu}$ and~$c \in K^*_{\mu}$, then~$\langle \lambda, c \rangle \geq 0$, where~$\langle \cdot, \cdot \rangle$ is the canonical scalar product.
It is worth noting that the relation $ K^*_{\mu}  =  K_{1/\mu}$ stands for second-order cones.

Eq.\,\eqref{eq:NCP} allows defining, for each contact $i\in \llbracket 1, n_c \rrbracket$, the primal and dual residuals as  \mbox{$\epsilon_{\text{p}}^{(i)} = \mathrm{dist}_{K_{\mu^{(i)}}}\left (\lambda^{(i)} \right )$} 
and \mbox{$\epsilon_{\text{d}}^{(i)} = \mathrm{dist}_{K_{\mu^{(i)}}^*}\left ( c^{(i)} + \Gamma\left(c^{(i)}, \mu^{(i)}\right) \right )$} respectively, where $\mathrm{dist}_\mathcal{C}$ is the distance function w.r.t. a convex set $\mathcal{C}$.
It also induces a contact complementarity criterion \mbox{$\epsilon_{\text{c}}^{(i)} =  \left |\langle \lambda^{(i)},  c^{(i)} + \Gamma\left(c^{(i)}, \mu^{(i)}\right) \rangle \right |$}.
From these per-contact criteria, it is then possible to introduce a well-posed absolute convergence criterion $\epsilon_{\text{abs}}$ for \eqref{eq:NCP}, as the maximum of $\epsilon_{\text{p}}^{(i)}$,$\epsilon_{\text{d}}^{(i)}$ and $\epsilon_{\text{c}}^{(i)}$ for all $i$.
We use this criterion as a stopping criterion in our implementation of NCP solvers, but also as a measure of physical accuracy in our experiments of Section~\ref{sec:exp}.
All the previous derivations were made with $\lambda$ being an impulse which causes it, and thus the criteria $\epsilon_{\text{p},\text{c}}$, to be proportional to the time step $\Delta t$.
However, it is preferable from the user-side to have $\epsilon_{\text{c}}$ and $\epsilon_{\text{p}}$ not correlated to $\Delta t$ so the precision threshold of the simulation $\epsilon_{\text{abs}}$ can be set independently of the time-step.
In practice, before solving we operate a change of variable to directly work on the equivalent contact forces $\frac{\lambda}{\Delta t}$. 
This is done by replacing $g$ and $c^*$ by their scaled counterpart $\frac{g}{\Delta t}$ and $\frac{c^*}{\Delta t}$ in the formulation of \eqref{eq:NCP}.
For readability purposes, equations are still written in impulse in what follows.

\begin{figure}
    \centering
    \includegraphics[width=1.\linewidth]{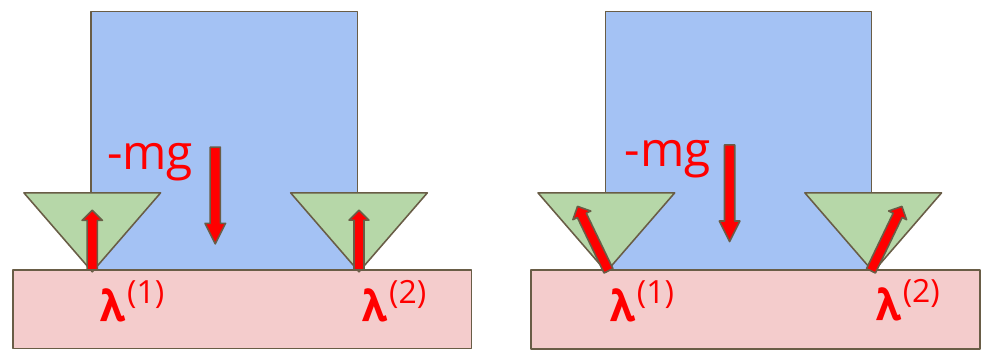}
    \caption{\textbf{Underdetermined contact problem.}
    The left and right contact forces are solutions of the NCP \eqref{eq:NCP} and lead to the same system velocity.
    Such an undetermined problem can also occur on normal forces.
    }
    \label{fig:underdetermined_contact_scheme}
    \vspace{-0.0cm}
\end{figure}

At this point, it is worth mentioning that the problem \eqref{eq:NCP}, which we refer to as NCP, does not derive from a convex optimization problem, thus making its solving complex.
Alternatively, one can see the frictional contact problem as two interleaved convex optimization problems~\cite{barbosa1987numerical, tzaferopoulos1993efficient, kaufman2008staggered,erleben2017prox,lelidec2021differentiable} whose unknowns, $\lambda$ and $v$, appear in both. 
Other formulations exist and we refer to \cite{acary2017contact} for a more complete review on the NCP.
Practically, the non-convexity can induce the existence of multiple, or even an infinite number of contact forces satisfying \eqref{eq:NCP}.
As mentioned earlier, this can be due to normal forces, but tangential components can also cause under-determination (Fig.~\ref{fig:underdetermined_contact_scheme}).
In this situation, it would be preferable for a simulator to provide the minimum norm solution in forces.
This property can prevent a simulator from exhibiting internal friction forces compressing or stretching the objects (Fig.~\ref{fig:underdetermined_contact_scheme}, right).
Indeed, such forces may not coincide with the forces observed by force sensors and would rather correspond to some internal deformations of the objects, which should thus be considered soft and no more rigid.
In the following, we will use the term “internal forces” to denote the force component deviating from the minimum norm solution.
Additionally, these internal forces might also be problematic as it is difficult to characterize them.
This may induce inconsistent derivatives, which become critical in the context of differentiable simulation. 

\begin{figure}[t!]
    \centering
    \includegraphics[width=.6\linewidth]{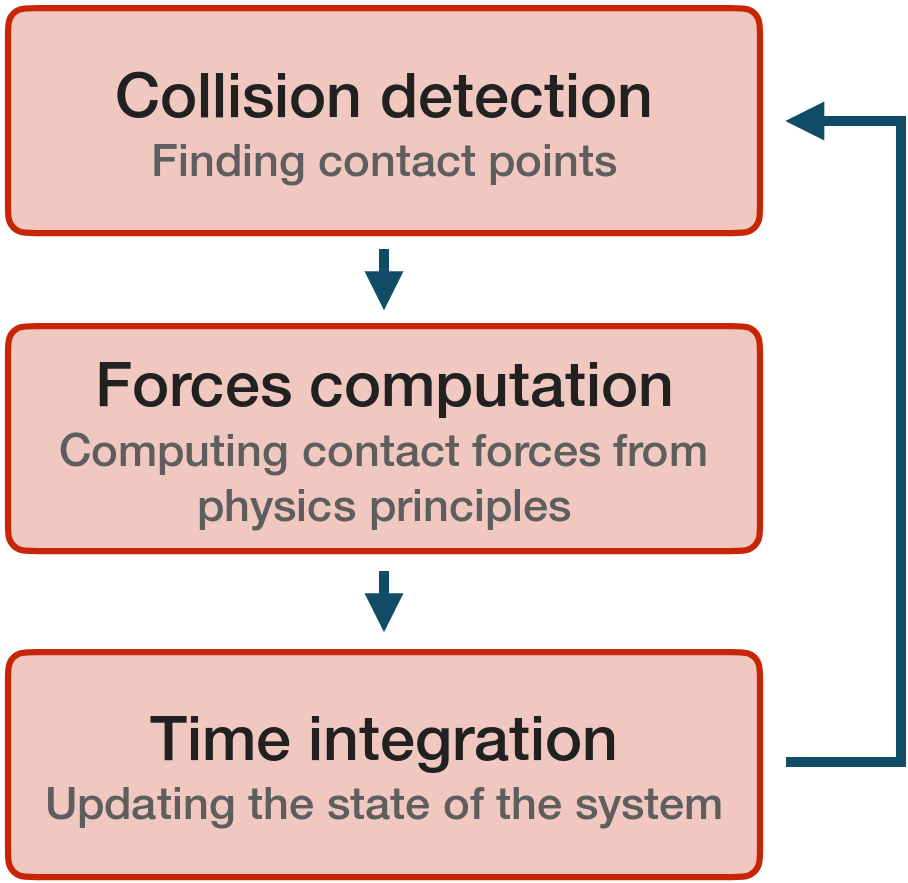}
    \caption{\textbf{Simulation routines.}
    When simulating rigid bodies with frictional contacts, a physics engine goes through a sequence of potentially challenging sub-problems: collision detection, contact forces computation, and integration time step.}
    \label{fig:simulator_structure}
    \vspace{-0.5cm}
\end{figure}

\noindent
\textbf{\\ Open-source frameworks for contact simulation.} 
To conclude this section, we propose to review the open-source software that is popular in the robotics community and that can be used for simulating contact.
Simulating contact interactions, as illustrated in Fig.~\ref{fig:simulator_structure}, involves two main stages, corresponding to the collision detection step (which objects are in contact) and the collision resolution (which contact forces are applied through the contact interaction).
These frameworks are enumerated in Tab.~\ref{tab:open_source_tools}.

More precisely, at each time step, a simulator must first detect which geometries are colliding and compute their separation vector $\Phi$.
The GJK and EPA algorithms are widely adopted for their low computational cost.
HPP-FCL\,\cite{hppfclweb,mirabel2016hpp,montaut2022collacc}, an extension of the Flexible Collision Library (FCL)\,\cite{panFCLGeneralPurpose2012} and libccd\,\cite{libccd} implement them efficiently.
Some simulators such as Bullet\,\cite{coumans2021}, ODE\,\cite{ode:2008} or PhysX~\cite{physx} also re-implement the same algorithm as an internal routine.

Once collisions are evaluated, one still requires the contact points free velocity $v_f$ and Jacobians $J$ to formulate \eqref{eq:NCP}.
These two quantities are efficiently computed via rigid body algorithms \cite{featherstone2014rigid}.
The RBDL\,\cite{felis2017rbdl} or the Pinocchio library\,\cite{carpentier2019pinocchio} provide efficient implementations to evaluate them.
In addition, Pinocchio proposes a direct and robust way to compute the Cholesky decomposition of the Delassus matrix G~\cite{carpentier2021proximal}.
These algorithms are also embedded as internal routines in various simulators such as MuJoCo\,\cite{todorov2012mujoco}, DART\,\cite{DART}, Drake\,\cite{drake},  Bullet\,\cite{coumans2021} or ODE\,\cite{ode:2008}, but they often are only partially exposed to the user.

Eventually, when all quantities necessary to formulate the NCP\,\eqref{eq:NCP} are computed, the simulator has to call a solver.
Every simulator, \textit{i.e.} MuJoCo\,\cite{todorov2012mujoco}, DART\,\cite{DART}, Bullet\,\cite{coumans2021}, Drake\,\cite{drake} and ODE\,\cite{ode:2008}, proposes its own implementation.
This procedure varies greatly depending on the physics engine, as each has its own physical and numerical choices.
In the next section, we detail the existing algorithms.

\begin{table}
  \caption{Open source tools for physics simulation in robotics.}
  \label{tab:open_source_tools}
  \centering
    \begin{tabular}{r|ccc}
      \toprule
                                       & License  & API & Used by  \\
      \midrule
      \multicolumn{4}{c}{\normalsize\textbf{Collision detection}}          \\
       FCL \cite{panFCLGeneralPurpose2012}   & BSD &  C++   &    DART, Drake   \\ \rowcolor{pastelblue}
       libccd \cite{libccd}                        & BSD &  C++, Python   &  \begin{tabular}{@{}c@{}}MuJoCo, Drake, \\ FCL, Bullet, ODE \end{tabular}      \\ 
      
       HPP-FCL \cite{hppfclweb}      & BSD &  C++, Python   &   Pinocchio   \\ \rowcolor{pastelblue}
       Bullet \cite{coumans2021}      & BSD &  C++, Python   &  DART    \\
       ODE \cite{ode:2008}      & BSD/GPL &  C++, Python   &  DART  \\
       \rowcolor{pastelblue}
       PhysX \cite{physx}      & BSD 3 &  C++, Python   &    \\
       \midrule
      \multicolumn{4}{c}{\normalsize\textbf{Rigid body dynamics algorithms}}                 \\
      Pinocchio \cite{carpentier2019pinocchio}                      & BSD &  C++, Python   &        \\
      \rowcolor{pastelblue}
      RBDL \cite{felis2017rbdl}   & zlib  &   C++, Python  &        \\
      Drake \cite{drake}  & BSD3  &   C++, Python  &        \\
      \midrule
      \multicolumn{4}{c}{\normalsize\textbf{Forces computation}}                 \\
      MuJoCo \cite{todorov2012mujoco}   & Apache 2.0  &   C++, Python  &        \\
      \rowcolor{pastelblue}
      DART \cite{DART}   & BSD 2  &   C++, Python  &        \\
      Bullet \cite{coumans2021}      & BSD &  C++, Python   &      \\
      \rowcolor{pastelblue}
      Drake \cite{drake}      & BSD 3 &  C++, Python   &      \\
      ODE \cite{drake}      & BSD/GPL &  C++, Python   &      \\
      \rowcolor{pastelblue}
      PhysX \cite{physx}      & BSD 3 &  C++, Python   &      \\
      \bottomrule
    \end{tabular}
  \vspace{-0.0cm}
\end{table}
\section{Algorithmic variations of the contact problem}  \label{sec:solvers}

\begin{table}
    \caption{Characteristics of various contact models.}
    \label{tab:contact_models}
    \centering
      \begin{tabular}{r|ccc}
        \toprule
                                           & Signorini & Coulomb & MDP \\
        \midrule
        \multicolumn{1}{l}{\normalsize\textbf{LCP}}         & \cmark &  &   \\
        
        \rowcolor{pastelblue}
        \multicolumn{1}{l}{\normalsize\textbf{CCP}}  &  &  \cmark   &  \cmark    \\

        \multicolumn{1}{l}{\normalsize\textbf{RaiSim} \cite{raisim}}                        &  \cmark   &  \cmark     &  \\
  
        \rowcolor{pastelblue}
        \multicolumn{1}{l}{\normalsize\textbf{NCP}}           &  \cmark &   \cmark  &  \cmark \\
        
        \bottomrule
      \end{tabular}
    \vspace{-0.0cm}
  \end{table}

\begin{table*}
  \caption{Characteristics of numerical algorithms.}
  \label{tab:contact_models_algorithms}
  \centering
    \begin{tabular}{r|cccc}
      \toprule
                                         &  Hard contacts & No internal forces & Robust  & Convergence guarantees\\
      \midrule
      \multicolumn{1}{l}{\normalsize\textbf{LCP}}              &   &    &   & \\
      \rowcolor{pastelblue}
      PGS \cite{coumans2021,ode:2008, physx, DART}         &  \cmark &    &  &  \\
      Staggered projections \cite{kaufman2008staggered}     &  \cmark  &  \cmark  & \cmark  & \\
      \midrule
      \multicolumn{1}{l}{\normalsize\textbf{CCP}}         &   &     & & \\
      \rowcolor{pastelblue}
       PGS \cite{anitescu2010iterative}                      &  \cmark&     &   & \cmark\\
      MuJoCo  \cite{todorov2012mujoco}                      &   & \cmark    & \cmark  & \cmark\\
      \rowcolor{pastelblue}
      ADMM (Alg.~\ref{alg:ccpadmm})                      &  \cmark & \cmark    & \cmark  & \cmark\\
      Drake \cite{castro2022unconstrained}               &   & \cmark    & \cmark  & \cmark\\
      \midrule
      \multicolumn{1}{l}{\normalsize\textbf{RaiSim} \cite{raisim}}                       &  \cmark &     & & \\

      \midrule
      \multicolumn{1}{l}{\normalsize\textbf{NCP}}                                  &   &    & & \\
      
      \rowcolor{pastelblue}
      PGS                         &  \cmark &      & &  \\
      
      Staggered projections \cite{lelidec2021differentiable}                        &  \cmark & \cmark    & \cmark  & \\

      \bottomrule
    \end{tabular}
  \vspace{-0.0cm}
\end{table*}

As explained in the previous section, the nonlinear complementarity problem\,\eqref{eq:NCP} does not derive from a variational principle but can be formulated as variational inequalities \cite{acary2017contact}. 
Thus, classical numerical optimization solvers cannot be used straightforwardly to solve it.
This section studies the various approximations and algorithmic techniques in the literature to tackle this problem.
{As summarized in Tab.~\ref{tab:contact_models_algorithms}, this section is organized into subsections describing the four contact models most commonly used in robotics, namely the linear complementarity problem (LCP), the cone complementary problem (CCP), RaiSim, and the nonlinear complementarity problem (NCP). For each contact model, we also report the related algorithmic variants.
If each tick in Tab.~\ref{tab:contact_models_algorithms} represents a positive point for the concerned algorithm, Sec.~\ref{sec:exp} shows that even one missing tick may be prohibitive and can cause a solver to be unusable in practice.
Finally, we also mention a set of useful implementation tricks that can be used to build an efficient simulator.}

\subsection{Linear Complementarity Problem}
A first way to simplify the solving of problem~\eqref{eq:NCP} is to linearize the NCP problem by approximating the second-order cone constraint from Coulomb's law with a pyramid, typically composed of four facets.
This is done by replacing $K_{\mu^{(i)}}$ by \mbox{$\Tilde{K}_{\mu^{(i)}} = \left\{\lambda | \lambda_N \geq 0, \|\lambda_T\|_\infty \leq \mu^{(i)} \lambda_N \right\}$}.
Doing so allows retrieving a linear complementarity problem (LCP), often easier to solve~\cite{cottle2009lcp}. 
Such a problem is more standard and better-studied than its nonlinear counterpart as it already has a long history of applications to frictional contacts~\cite{mitsopoulou1987contribution,doudoumis1988solution,jourdan1998gauss,erleben2015LCP}.

\noindent
\textbf{\\Direct methods} for LCP date back to the 1960s and are available options in well-known simulators such as ODE \cite{ode:2008} and Bullet \cite{coumans2021} which implement respectively the Lemke's~\cite{lemke1965bimatrix} and Dantzig's~\cite{dantzig1963positive} algorithms. 
Under specific circumstances \cite{anitescu1997formulating}, the algorithm is guaranteed to find a solution.

\noindent
\textbf{\\Projected Gauss-Seidel.} Due to its easy implementation and the possibility to early-stop it, the projected Gauss-Seidel (PGS) algorithm\,(Alg.~\ref{alg:lcppgs}) algorithm represents an attractive alternative and was widely adopted as the default solver by many physics engines, such as in Bullet \cite{coumans2021}, PhysX \cite{physx}, ODE \cite{ode:2008}, and DART \cite{DART,werling2021fast} simulators.
This iterative algorithm loops on contact points and successively updates the normal and tangent contact forces.
Because PGS works separately on each contact point, the update compensates for the current errors due to the estimated forces from other contact points.
Yet, as illustrated in the experimental section~\ref{sec:exp}, this process induces the emergence of internal forces during the solving. 
Moreover, Gauss-Seidel-based approaches are similar to what is also known as block coordinate descent in the optimization literature. As first-order algorithms, they do not benefit from improved convergence rates or robustness with respect to their conditioning, unlike second-order algorithms.

\begin{figure}
    \centering
    \includegraphics[width=.45\linewidth]{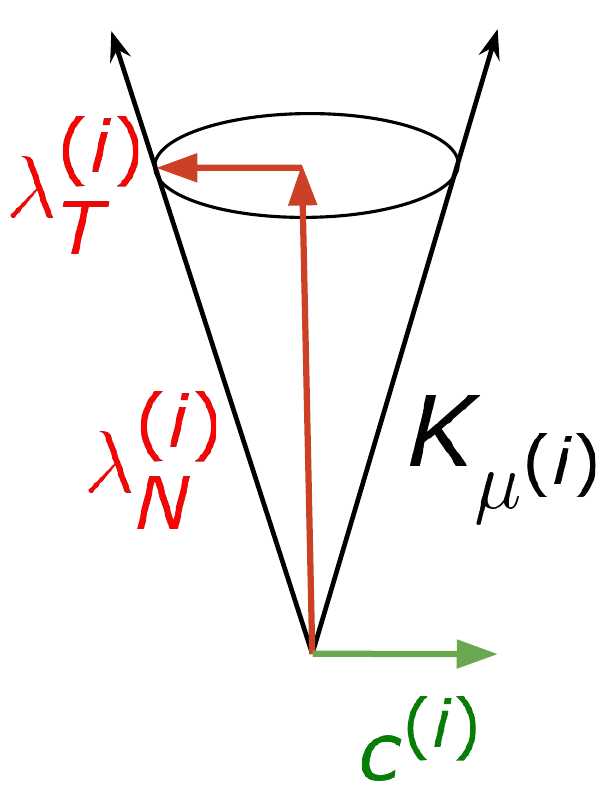}
    \includegraphics[width=.45\linewidth]{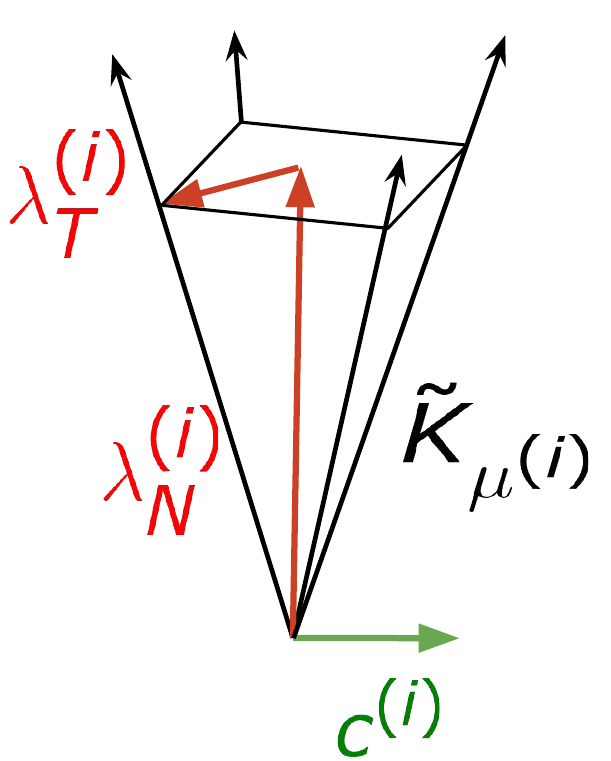}
    \caption{\textbf{Cone linear approximation.} 
    Linearizing the friction cone induces a bias in the direction of friction forces. 
    The MDP tends to push tangential forces toward the corners of the pyramid.}
    \label{fig:lcp_bias}
    \vspace{-0.0cm}
\end{figure}

In parallel, the linearization of the second-order cone causes the loss of the isotropy for friction, as stated by Coulomb's law.
By choosing the axes for the facets of the pyramid and due to the maximum dissipation principle, it is established that one incidentally biases the friction forces towards the corners \cite{trinkle1997dynamic,acary2008numerical}, as illustrated in Fig.~\ref{fig:lcp_bias}.
This error is sometimes mitigated by increasing the number of facets, which also comes at the cost of more computations.\\
\begin{algorithm}[t]
    \SetAlgoLined
    \KwIn{Delassus matrix: $G$, free velocity: $g$, friction cones: $K_\mu$}
    \KwOut{Contact forces: $ \lambda$}
    \For{$k=1$ \KwTo $n_{iter}$}{
       \For{$i=1$ \KwTo $n_c$}{
       $\lambda^{(i)}_N \leftarrow \lambda^{(i)}_N - \frac{1}{G^{(ii)}_{NN}}\left(G\lambda+g \right)^{(i)}_N$; \label{line:lcppgs_normal}\\
       $\lambda^{(i)}_N \leftarrow \max (0, \lambda^{(i)}_N)$;\\
       $\lambda^{(i)}_T \leftarrow \lambda^{(i)}_T - \frac{1}{\min (G^{(ii)}_{T_x T_x},G^{(ii)}_{T_y T_y})}\left(G\lambda+g \right)^{(i)}_T$; \label{line:lcppgs_tangent}\\
       $\lambda^{(i)}_T \leftarrow \text{clamp}(\lambda^{(i)}_T, \mu_i \lambda_N)$;\label{line:clamping}
       
    }
    }
    \caption{Pseudocode of the projected Gauss-Seidel (PGS) algorithm for solving LCPs.}
    \label{alg:lcppgs}
\end{algorithm}

\subsection{Cone Complementarity Problem.}\label{subsec:ccp}
An alternative approach consists of approximating the NCP problem in order to transform it into a more classical convex optimization problem.
By relaxing the complementarity constraint from \eqref{eq:NCP}, one can obtain a Cone Complementarity Problem (CCP) \cite{anitescu2006optimization}:
\begin{equation}
\label{eq:CCP}
    K_\mu \ni \lambda \perp c \in K_\mu^*
\end{equation}
If this relaxation preserves the Maximum Dissipation Principle (MDP) and the second-order friction cone, it loses the \textit{Signorini condition} \eqref{eq:signorini_cond}.
Indeed, re-writing the complementarity of \eqref{eq:CCP} yields:
\begin{equation}
    \forall i , \ {\lambda_N^{(i)} } c_N^{(i)} + {\lambda_T^{(i)}}^\top c_T^{(i)} = 0,
\end{equation}
and if the i\textsuperscript{th} contact point is sliding, the MDP \eqref{eq:mdp} leads to:
\begin{equation}
    {\lambda_N^{(i)}}  c_N^{(i)} -  \mu^{(i)} \lambda_N^{(i)} \|c_T^{(i)}\|_2 = 0,
\end{equation}
which is equivalent to the following complementarity condition:
\begin{equation}
    \label{eq:ccp_comp}
    \lambda_N^{(i)} \perp (c_N^{(i)} - \mu \|c_T^{(i)}\|_2) .
\end{equation}
\eqref{eq:ccp_comp} indicates that the CCP approximation allows for simultaneous normal velocity and forces, contrary to \eqref{eq:signorini_cond}.
In practice, this results in objects interacting at distance when contact points are sliding.
In its seminal work \cite{anitescu2006optimization}, Anitescu shows the interaction distance to be $\Delta t \mu \|c_T^{(i)}\|$.
It is worth insisting on the fact that such an artifact only emerges in the case of a sliding contact and can be mitigated, and even controlled, with smaller time steps and sliding velocities. 
Moreover, it is still under debate to determine if this behavior is prohibitive for robotics applications.

Because CCP \eqref{eq:CCP} approximates the NCP \eqref{eq:NCP}, the convergence is checked via a different criterion.
In fact, in the same way, the De Saxc\'e correction was ignored in \eqref{eq:CCP}, a convergence criterion is obtained by removing this term from the dual convergence criterion $\epsilon_{\text{d}}$ of the NCP introduced previously.

\noindent
\textbf{\\PGS.} The PGS algorithm can be directly adapted to handle the CCP problem \cite{anitescu2010iterative} (Alg.~\ref{alg:ccppgs}) but inherits from the first-order convergence rates (\cite{anitescu2010iterative} exhibits in the order of hundreds of iterations to converge in general).
In the light of what follows, the algorithm even becomes equivalent to a projected gradient descent which is a classical constrained optimization technique.

\SetKwRepeat{Do}{do}{while}
\begin{algorithm}[t]
    \SetAlgoLined
    \KwIn{Delassus matrix: $G$, free velocity: $g$, friction cones: $K_\mu$}
    \KwOut{Contact forces: $ \lambda$}
    \For{$k=1$ \KwTo $n_{iter}$}{
       \For{$i=1$ \KwTo $n_c$}{
       $\lambda^{(i)} \leftarrow \lambda^{(i)} - \frac{3}{G^{(ii)}_{NN}+G^{(ii)}_{T_x T_x}+G^{(ii)}_{T_y T_y}}\left(G\lambda+g \right)^{(i)}$;\label{line:ccppgs_block_inverse}\\
       $\lambda^{(i)} \leftarrow \text{proj}_{K_{\mu_i}}(\lambda^{(i)})$; \label{line:normal_proj}\\
    }
    }
    \caption{Projected Gauss-Seidel (PGS) algorithm for the dual Cone Complementarity Problem (CCP)}
    \label{alg:ccppgs}
\end{algorithm}

\noindent
\textbf{\\Optimization on the dual.} The problem \eqref{eq:CCP} can, in fact, be viewed as the Karush-Kuhn-Tucker conditions of an equivalent Quadratically Constrained Quadratic Programming (QCQP) problem:
\begin{equation}
\label{eq:CCP_QCQP}
    \min_{\lambda \in K_\mu} \frac{1}{2}\lambda^\top G \lambda + g^\top \lambda
\end{equation}
Once the contact problem is formulated as an optimization problem, any optimization algorithms can be employed to solve it and classical optimization theory provides convergence guarantees.
Here, we propose to study an ADMM algorithm \cite{boyd2011distributed}, an advanced first-order algorithm known to be efficient to reach mild accuracy and which can stall when further improving the solution.
As pointed out in  \cite{todorov2012mujoco}, Interior Point algorithms \cite{mehrotra1992implementation} could also be used to reach higher precision solutions even though we do not find them in any of the robotics simulators here
mentioned.
A benefit of using the family of proximal algorithms like ADMM is their natural ability to handle the numerical issues coming from ill-conditioned and hyper-static cases \cite{nocedal1999numerical,parikh2014proximal}.
This property makes it possible to accurately simulate hard contacts, \textit{i.e.}, without any shift due to compliance $R=0$, and is reported in Tab.~\ref{tab:contact_models_algorithms} by the "hard contacts" column. 
Another by-product of such methods is the implicit regularization they induce on the found solution, which removes the potential internal forces. 
This last property is an empirical observation resulting from the experimental section~\ref{sec:exp} and, to our knowledge, has not yet been proven by the literature of proximal optimization. 
Therefore, it remains to be confirmed by subsequent work.

One may argue that such algorithms require to compute $G^{-1}$ (Alg.~\ref{alg:ccpadmm}, line~\ref{line:primal1_update}) while per contact approaches repeatedly solve for each contact point individually, and thus only require the cheap inverse of diagonal blocks from $G$ (Alg.~\ref{alg:lcppgs}, lines~\ref{line:lcppgs_normal},\ref{line:lcppgs_tangent}, Alg.~\ref{alg:ccppgs}, line~\ref{line:ccppgs_block_inverse}, Alg.~\ref{alg:raisim}, line~\ref{line:raisim_block_inverse}, Alg.~\ref{alg:ncppgs}, lines~\ref{line:ncppgs_normal},\ref{line:ncppgs_tangent}).
However, the recent progress \cite{carpentier2021proximal} demonstrated the Cholesky decomposition of $G$ can be computed efficiently and robustly.
We detail this point later when discussing implementation tricks, at the end of this section.
Exploiting the knowledge of $G^{-1}$ and not the block components as in the "per-contact" approaches mentioned earlier allows us to capture the coupling between all contact points.
\begin{algorithm}[t]
    \SetAlgoLined
    \KwIn{Delassus matrix: $G$, free velocity: $g$, friction cones: $K_\mu$}
    \KwOut{Contact forces: $ \lambda$}
    $\Tilde{G}^{-1} \leftarrow (G + \rho \mathrm{Id})^{-1}$ \label{line:reg_inverse}\;
    \For{$k=1$ \KwTo $n_{iter}$}{
        $\lambda \leftarrow -\Tilde{G}^{-1}(g-\rho z + \gamma)$; \label{line:primal1_update}\\
        $ z\leftarrow \text{proj}_{K_\mu}(\lambda + \frac{\gamma}{\rho})$;\\
        $\gamma \leftarrow \gamma + \rho (\lambda - z)$;\\
    }
    \caption{ADMM algorithm for the dual Cone Complementarity Problem (CCP)}
    \label{alg:ccpadmm}
\end{algorithm}

\noindent
\textbf{\\Optimization on the primal.} By reverse engineering, it is possible to form an optimization problem on joint velocities $v$ whose dual would be \eqref{eq:CCP_QCQP}. 
This approach is adopted in both MuJoCo\cite{todorov2012mujoco} and Drake\cite{castro2022unconstrained} and results in the following optimization problem:
\begin{align}
    \label{eq:ccp_primal}
    \min_{v,y} & \ \frac{1}{2} \big \| v - v^f \big \|_M^2 + \frac{1}{2} \|y - c^* \|^2_{R^{-1}}  \\
    & s.t. \ Jv  - y \in K^*_\mu \nonumber
\end{align}
where $\| x \|_X = \sqrt{x^\top X x}$ with $X\succ 0$.
Working on the equations, this problem can be formulated as an unconstrained optimization problem:
\begin{align}
    \label{eq:ccp_primal_unconstrained}
    \min_{v} \ l_p(v) =  \frac{1}{2} \big \| v - v^f \big \|_M^2 + \frac{1}{2} \|\mathcal{P}^R_{\mathcal{K}_\mu}(y(Jv))\|_R^2
\end{align}
where $\mathcal{P}^R_{\mathcal{K}_\mu}(y) = \argmin_{\gamma \in \mathcal{K}_\mu}  \|\gamma - y \|_R^2 $; $y(c) =  - R^{-1} ( c - c^*)$; and which is viable only when $R$ is non-null.
The latter condition makes it impossible to model hard contacts.
As evoked earlier, this is equivalent to replacing $G$ by $\Tilde{G} = G + R$ in the quadratic part of \eqref{eq:CCP_QCQP}, which is justified by a compliant contact hypothesis.
Indeed, $R$ corresponds to a compliance, which should be a material property of the objects involved in the collision.
However, MuJoCo arbitrarily sets this to the diagonal of $\alpha G$, where $\alpha \in [ 0,1]$ is close to $0$.
This choice has no physical justification (at least, without making strong assumptions that are not met in practice), and its only intent is to improve the conditioning of the problem to ease the solving and artificially make the solution unique.
Moreover, $R$ has non-null tangential components and thus may also introduce some tangential "compliance" which corresponds to the relaxation of Coulomb's law (Fig.~\ref{fig:sig_coulomb}).
In fact, this should instead be interpreted as a Tikhonov regularization term enforcing the strict convexity of the problem to facilitate the numerics and the existence of both the forward and inverse dynamics computation at the cost of shifting, even more, the solution.
Drake's algorithm \cite{castro2022unconstrained} improves this point by providing a more physical way of setting $R$.

Both Drake and MuJoCo use a Newton solver to tackle \eqref{eq:ccp_primal_unconstrained} (Alg.~\ref{alg:ccpnewton}).
Due to the non-linearity of the second term of \eqref{eq:ccp_primal_unconstrained}, this approach requires updating the inverse of the Hessian at every iteration (Alg.~\ref{alg:ccpnewton},line~\ref{line:ccpnewton_hessian}). 
As proposed in \cite{castro2022unconstrained}, the use of advanced algebra routines allows to reduce the computational burden of each step.
In this work, we provide an implementation of the Newton algorithm with an Armijo backtracking line search (using parameters from \cite{castro2022unconstrained}).
MuJoCo and Drake additionally implement an exact line search which improves performance and this difference should be kept in mind when interpreting the results obtained with our implementation.

\begin{algorithm}[t]
    \SetAlgoLined
    \KwIn{Inertia matrix: $M$, Jacobian of contacts: $J$, compliance: $R$, free velocity: $v^f$, friction cones: $K_\mu$}
    \KwOut{Joint velocity: $v$}
    \For{$k=1$ \KwTo $n_{iter}$}{
        $\nabla_v l_p \leftarrow M (v- v^f) - J^\top \mathcal{P}^R_{\mathcal{K}_\mu}(y(Jv))$; \label{line:ccpnewton_grad}\\
       $ H \leftarrow M + J^\top \nabla_v \mathcal{P}^R_{\mathcal{K}_\mu}(y(Jv)) J$ ;\label{line:ccpnewton_hessian}\\
        $\Delta v \leftarrow - H^{-1} \nabla_v l_p$ \label{line:ccpnewton_descentdir};\\
        $\alpha \leftarrow \argmin_\beta l_p(v+\beta \Delta v)$ \label{line:ccpnewton_linesearch};\\
        $v \leftarrow v + \alpha \Delta v$ \label{line:ccpnewton_step};\\
    }
    \caption{Newton algorithm for the primal Cone Complementarity Problem (CCP)}
    \label{alg:ccpnewton}
\end{algorithm}

\subsection{Raisim contact model}

A contact model introduced in \cite{preclik2014models} and implemented in the RaiSim simulator~\cite{raisim} aims at partially correcting the drawbacks from the CCP contact model exploited in MuJoCo~\cite{todorov2012mujoco} and Drake~\cite{drake}.
As explained earlier, the CCP formulation relaxes the Signorini condition for sliding contacts, leading to positive power from normal contact forces.
The contact model proposed in \cite{preclik2014models} fixes this by explicitly enforcing the Signorini condition by constraining $\lambda^{(i)}$ to remain in the null normal velocity hyper-plane $ V_N^{(i)} = \{\lambda | G_N^{(ii)} \lambda + \Tilde{g}_N^{(i)} = 0 \}$ where $\Tilde{g}^{(i)} = g^{(i)} +\sum_{j\neq i} G^{(ij)}\lambda^{(j)}$ is the i\textsuperscript{th} contact point velocity as if it were free.
Here, we generalize the use of the subscript and the superscript introduced previously to matrices, where a second superscript (or subscript) corresponds to a slicing operation on the columns \textit{e.g.} $G^{(ij)} \in \mathbb{R}^{3\times 3}$ denotes the sub-block of $G$ whose rows are associated to the i\textsuperscript{th} contact and columns to the j\textsuperscript{th} contact.
For a sliding contact point, the problem \eqref{eq:CCP_QCQP} becomes:
\begin{equation}
\label{eq:Raisim_QCQP}
    \min_{\lambda \in K_{\mu^{(i)}} \cap V_N^{(i)}} \frac{1}{2}\lambda^\top G^{(i)} \lambda + {\Tilde{g}^{(i)\top}} \lambda
\end{equation}
The new problem \eqref{eq:Raisim_QCQP} remains a QCQP and \cite{raisim} leverages the analytical formula of the ellipse $K_{\mu^{(i)}} \cap V_N^{(i)}$ in polar coordinates to tackle it as a 1D problem via the bisection algorithm \cite{boyd2007localization} (Alg.~\ref{alg:raisim}, line~\ref{line:bisection}).
We refer to the original publication for a more detailed description of the bisection routine \cite{raisim}.
\begin{figure}
    \centering
    \includegraphics[width=.55\linewidth]{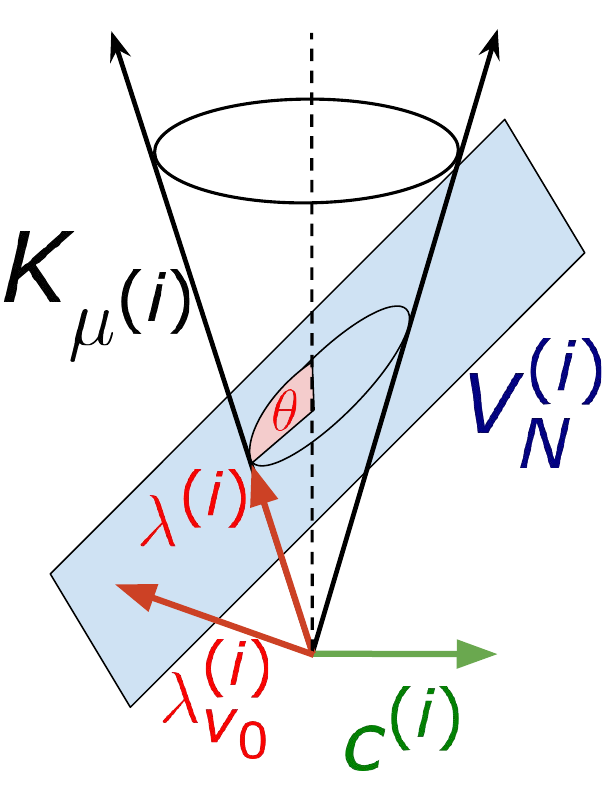}
    \caption{\textbf{Bisection algorithm.} 
    When the contact point is sliding, $\lambda_{v_0}^{(i)}$ (Alg.~\ref{alg:raisim}, line~\ref{line:raisim_block_inverse}) lies outside the friction cone $K_{\mu^{(i)}}$, leading to a non-null tangential contact velocity $c^{(i)}$.
    In this case, RaiSim solves \eqref{eq:Raisim_QCQP}.
    This is equivalent to finding the \mbox{$\lambda \in K_{\mu^{(i)}} \cap V_N^{(i)}$} which is the closest to $\lambda_{v_0}^{(i)}$ under the metric defined by $G^{(ii)}$.
    The constraint set being an ellipse, the problem boils down to a 1D problem on $\theta$ using polar coordinates.
    This figure is inspired from Fig.~2 of \cite{raisim}.
    }
    \label{fig:bisection_algo}
    \vspace{-0.0cm}
\end{figure}

This approach implies several drawbacks.
Indeed, it requires knowing whether a contact point is sliding, which cannot be known in advance as the contact point velocity depends on the contact forces.
Thus, some heuristics, based on the disjunctive formulation of the contact problem \eqref{eq:disjunctive_form}, are introduced to try to guess the type of contact which will occur, \textit{i.e.} take-off (Alg.~\ref{alg:raisim}, line~\ref{line:takeoff}), sticking (Alg.~\ref{alg:raisim}, line~\ref{line:stiction}) or sliding (Alg.~\ref{alg:raisim}, line~\ref{line:sliding}). 
Such heuristics may be wrong, which may cause the algorithms to get stuck and lose convergence guarantees.
This effect is strengthened by the caveats of the per-contact loop, which additionally make RaiSim not robust to conditioning and prone to internal forces.
Eventually, if adding the constraint $\lambda^{(i)} \in V_N^{(i)}$ allows retrieving the Signorini condition from the CCP model, it also induces the loss of the maximum dissipation principle.
Writing the Karush Kuhn Tucker\,(KKT) conditions of the problem \eqref{eq:Raisim_QCQP} and some algebra manipulations yields:
\begin{equation}
\label{eq:Raisim_kkt}
    c_T^{(i)} \propto - \lambda_T^{(i)} - \frac{{\mu^{(i)}}^2\lambda_N^{(i)}}{G_{NN}^{(ii)}}G_{NT}^{(ii)}
\end{equation}
which contradicts \eqref{eq:mdp}.

The problem solved by RaiSim depends on the contact mode, \textit{e.g.} \eqref{eq:Raisim_QCQP} is solved only for a sliding contact and would require the computation of the unknown dual variable.
Therefore, it is more complex to define a proper convergence criterion than in previous cases \eqref{eq:NCP} and \eqref{eq:CCP}.
In this respect, either a fix-point criterion \textit{i.e} the distance between two consecutive iterates, or the previously defined NCP criterion \eqref{eq:NCP} can be used to coarsely monitor convergence.
We chose the latest in order to have a criterion homogeneous to the ones used for \eqref{eq:NCP} and \eqref{eq:CCP}.

\begin{algorithm}[t]
    \SetAlgoLined
    \KwIn{Delassus matrix: $G$, free velocity: $g$, friction cones: $K_\mu$}
    \KwOut{Contact forces: $ \lambda$, velocity: $v$}
    \For{$k=1$ \KwTo $n_{iter}$}{
       \For{$i=1$ \KwTo $n_c$}{
       $\Tilde{g}^{(i)} \leftarrow g^{(i)} + \sum_{j\neq i} G^{(ij)}\lambda^{(j)}$\;
       $\lambda_{v_0}^{(i)}\leftarrow - {G^{(ii)}}^{-1} \Tilde{g}^{(i)}$\; \label{line:raisim_block_inverse}
       \uIf{$\Tilde{g}^{(i)}_N>0$}{\label{line:takeoff}
  \tcp{takeoff}
    $\lambda^{*} \leftarrow 0$\;
  }
  \uElseIf{$\lambda_{v_0}^{(i)} \in K_{\mu^{(i)}}$}{\label{line:stiction}
       \tcp{stiction}
    $\lambda^{*} \leftarrow \lambda_{v_0}^{(i)}$\;
  }
  \Else{ \label{line:sliding}
    \tcp{sliding}
    $\lambda^{*} \leftarrow \mathrm{bisection}(G^{(ii)}, \Tilde{g}^{(i)}, K_{\mu^{(i)}},\lambda_{v_0}^{(i)})$\; \label{line:bisection}
  }
  $\lambda^{(i)} \leftarrow (1-\alpha) \lambda^{(i)} + \alpha \lambda^{*}$\;
  $\alpha \leftarrow \gamma \alpha + (1-\gamma) \alpha_{min}$ \;
  
    }
    }
    \caption{Per-contact bisection algorithm}
    \label{alg:raisim}
\end{algorithm}

\subsection{Tackling the NCP}

Despite the non-smooth and non-convex issues described previously, some simulation algorithms aim to directly solve the original NCP problem~\cite{jourdan1998gauss,acary2017contact,macklin2019non,howelllecleach2022}.
\noindent
\textbf{\\PGS.}
The PGS algorithm exploited for LCP and CCP problems can easily be adapted to the NCP case by changing the clamping step (Alg.~\ref{alg:lcppgs},line~\ref{line:clamping}) or the normal projection (Alg.~\ref{alg:ccppgs}, line~\ref{line:normal_proj}) for a horizontal projection on the cone (Alg.~\ref{alg:ncppgs}, line~\ref{line:horizontal_proj}).
However, it is worth noting that such approaches have fewer convergence guarantees than their relaxed counterpart \cite{acary2011contact}. 
As with every Gauss-Seidel approach, the methods inherited from the sensitivity to ill-conditioning and jamming internal forces.  
\begin{algorithm}[t]
    \SetAlgoLined
    \KwIn{Delassus matrix: $G$, free velocity: $g$, friction cones: $K_\mu$}
    \KwOut{Contact forces: $ \lambda$, velocity: $v$}
    \For{$k=1$ \KwTo $n_{iter}$}{
       \For{$i=1$ \KwTo $n_c$}{
       $\lambda^{(i)}_N \leftarrow \lambda^{(i)}_N - \frac{1}{G^{(ii)}_{NN}}\left(G\lambda+g \right)^{(i)}_N$;\label{line:ncppgs_normal}\\
       $\lambda^{(i)}_N \leftarrow \max (0, \lambda^{(i)}_N)$;\\
       $\lambda^{(i)}_T \leftarrow \lambda^{(i)}_T - \frac{1}{\min (G^{(ii)}_{T_x T_x},G^{(ii)}_{T_y T_y})}\left(G\lambda+g \right)^{(i)}_T$; \label{line:ncppgs_tangent}\\
       $\lambda^{(i)}_T \leftarrow proj_{\mu_i\lambda_N^{(i)}}(\lambda^{(i)}_T)$; \label{line:horizontal_proj}
    }
    }
    \caption{Projected Gauss-Seidel (PGS) algorithm for Non-linear Complementarity Problem (NCP)}
    \label{alg:ncppgs}
\end{algorithm}

\noindent
\textbf{\\Staggered-projections.}
The staggered projections  (Alg.~\ref{alg:stag_proj}) approach, appearing in \cite{barbosa1987numerical, tzaferopoulos1993efficient} and implemented in a simulator in \cite{kaufman2008staggered,lelidec2021differentiable}, proceeds by rewriting the NCP as two interleaved optimization problems.
This interconnection is solved via a fix-point algorithm that repeatedly injects one problem's solution into the formulation of the other.
The staggered projection algorithm has no convergence guarantees but was heavily tested and seems, in practice, to converge most of the time in a few iterations (typically five iterations~\cite{lelidec2021differentiable}).
Solving a cascade of optimization problems allows the use of robust optimization algorithms (\textit{e.g.}, ADMM), but remains more costly than other approaches. 
\begin{algorithm}[t]
    \SetAlgoLined
    \KwIn{Delassus matrix: $G$, free velocity: $g$, friction cones: $K_\mu$}
    \KwOut{Contact forces: $ \lambda$, velocity: $v$}
    \For{$k=1$ \KwTo $n_{iter}$}{
       $\Tilde{g_N} \leftarrow g_T + G_{NT}\lambda_T$;\\
       $\lambda_N \leftarrow \argmin_{\lambda \geq 0} \frac{1}{2}\lambda^\top G_N \lambda + \Tilde{g_N}^\top \lambda$; \\
       $\Tilde{g_T} \leftarrow g_T + G_{TN}\lambda_N$;\\
       $\lambda_T \leftarrow \argmin_{\|\lambda^{(i)}\| \leq \mu_i \lambda_N^{(i)}}\frac{1}{2}\lambda^\top G_T \lambda + \Tilde{g_T}^\top \lambda$;\\
    }
    \caption{Staggered projections algorithm}
    \label{alg:stag_proj}
\end{algorithm}

\begin{figure*}[t]
    \centering
    \includegraphics[width=.32\linewidth]{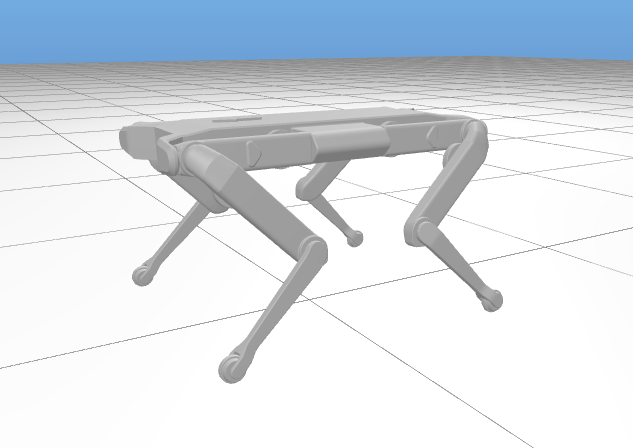}
    \includegraphics[width=.193\linewidth]{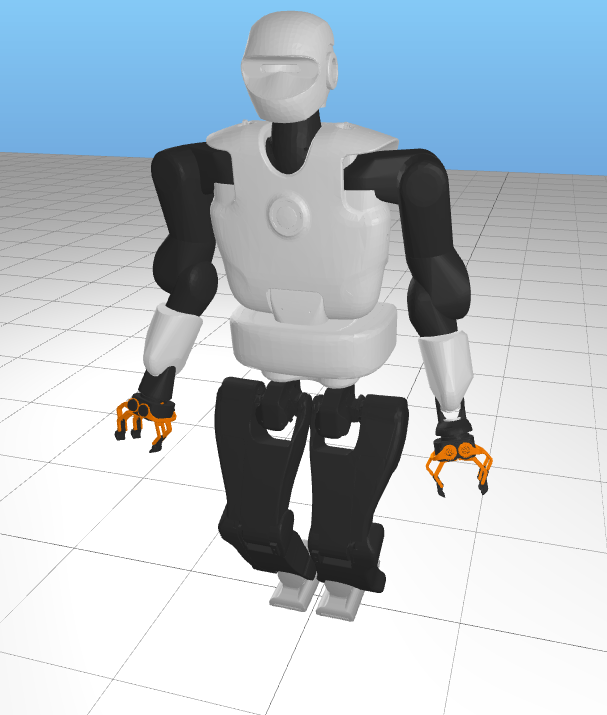}
    \includegraphics[width=.26\linewidth]{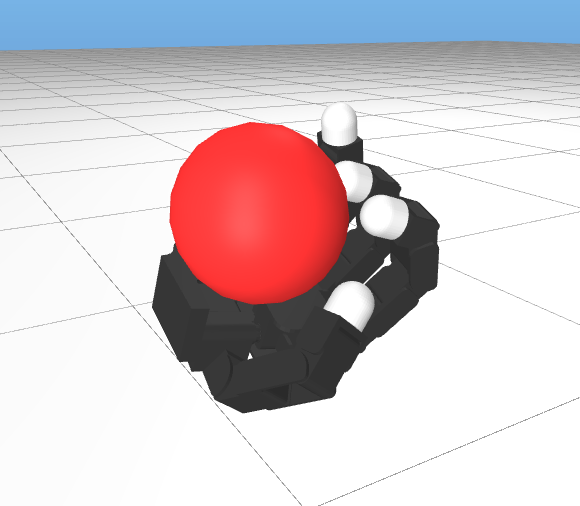}
    \caption{\textbf{Robotics systems used for the experiments}.
    The Solo-12 quadruped (\textbf{Left}), the Talos humanoid (\textbf{Center}), and the Allegro hand (\textbf{Right}) allow to respectively exhibit locomotion, high-dimensional, and manipulation contact scenario.}
    \label{fig:robot_exp}
    \vspace{-0.0cm}
\end{figure*}

\subsection{Implementation details}
\label{sec:implementation_details}

In practice, the performances of contact solvers can be improved by a few simple tricks. 

\noindent
\textbf{\\Warm-starting } the solver by providing the contact forces from the previous time step allows to greatly reduce the required computation.
Indeed, in the case of a persisting contact between two objects, the contact forces are being cached and reused as an initial guess when solving for the contact forces of the next time step.
This relies on the ability of the contact solver algorithm to be warm-started.
This excludes Interior Point \cite{mehrotra1992implementation} algorithms, as they would only benefit from an initial guess close to the so-called central path \cite{nocedal1999numerical}.
By contrast, the feasible set of contact forces may change from one time step to the other, even in the case of a persisting contact point.
On the opposite, ADMM and, more generally, Augmented Lagrangian (AL) methods can naturally be warm started: not only the primal (\textit{i.e.}, contact velocities) and dual (\textit{i.e.}, contact forces) variables, but also the proximal parameter is initialized with the previous values.

\noindent
\textbf{\\Cholesky computation.}
In addition, second-order algorithms can further exploit the recent progress in rigid body algorithms \cite{carpentier2021proximal}. 
This work takes advantage of the sparse structure of the kinematic chains in order to efficiently compute the Cholesky decomposition of the Delassus matrix $G$.
This approach is robust enough to handle the case of hyperstatic systems and reduces the cost of the computation of matrix-vector products involving $G^{-1}$ (Alg.~\ref{alg:ccpadmm}, line~\ref{line:primal1_update}).
This also indicates that evaluating $G$ from its Cholesky decomposition, as required by per-contact approaches, actually constitutes an additional cost.

\noindent
\textbf{\\Proximal parameter adaptation.}
In the context of ADMM (Alg.~\ref{alg:ccpadmm}), the algorithm from \cite{carpentier2021proximal} can also be favorably combined with the adaptation of the proximal parameter.
Indeed, updating the regularized Cholesky can be done at almost no cost by using \cite{carpentier2021proximal}.
In our implementation, we follow the work from \cite{simple} to detect when and how $\rho$ should be adapted.
More precisely, whenever the primal residual is significantly greater than the dual one (a threshold for the ratio has to be set, a typical value being 10), $\rho$ should be increased in order to better enforce the constraint of the problem and thus, reduce the primal residual.
The proximal parameter $\rho$ is then updated via a spectral rule which multiplies it by $\kappa^{0.05}$, $\kappa$ being the condition number of the Delassus matrix, defined as the ratio between the largest and the smallest eigenvalues.
Conversely, whenever the dual residual dominates the primal one, $\rho$ should be decreased by dividing it with the same factor.
The condition number $\kappa$ can be efficiently evaluated beforehand via a power-iteration algorithm whose iterates have a computational cost equivalent to the ones from the ADMM algorithm.
This procedure is detailed and evaluated in \cite{simple}.
Alternatively, we could use a linear update rule for $\rho$ as it is done in OSQP \cite{osqp} which would be less efficient in the case of ill-conditioned problems.

\noindent
\textbf{\\Over-relaxation.}
Additionally, over-relaxation is often employed to accelerate the convergence of both Gauss-Seidel and ADMM algorithms. 
This technique applies the following update:
\begin{equation}
    \lambda \leftarrow \alpha \lambda^- + (1-\alpha) \lambda,
\end{equation}
where $\alpha \in \left ] 0,2 \right [ $ and $\lambda^-$ denotes the previous iteration.
For $\alpha > 1$, over-relaxing consists of an extrapolation step and should be carefully used, as it may also hinder convergence.
Typically, setting $\alpha$ to $0.8$ improved convergence of the PGS algorithm.

\section{Experiments}
\label{sec:exp}

In this section, we evaluate the performances and behaviors of the formulations explained in Sec.~\ref{sec:solvers}.
To fairly compare and benchmark the various algorithmic formulations, we have implemented them in a unified C++ framework called ContactBench. 
In the following, we denote by RaiSim and Drake our re-implementation of the contact solvers described in the corresponding papers \cite{raisim,castro2022unconstrained}.
For Drake, it is worth noting that our implementation uses a backtracking line-search with an Armijo condition instead of the line-search proposed in the original paper \cite{castro2022unconstrained}.
Our framework extensively relies on the Pinocchio library \cite{carpentier2019pinocchio} for rigid body algorithms and HPP-FCL \cite{hppfclweb,montaut2022collision} implementation of GJK and EPA for collision detection.
Our code is made open-source in the Contactbench C++ library (\url{https://github.com/Simple-Robotics/contactbench}).

Several factors may hinder the correctness and accuracy of simulators based on time-stepping methods:
\begin{enumerate}[label=\roman*)]
    \item the low accuracy of the solver of the contact problem;
    \item the limitation from the contact model itself;
    \item or the numerical integration due to the time discretization scheme.
\end{enumerate}

In this section, we first evaluate the error from sources i) and ii) (Sec.~\ref{sec:exp_phys}).
The source of error i) is evaluated by measuring the time taken to reach a given accuracy.
The errors from ii) are analyzed by measuring the residual for an (approximately) infinite time budget.
We further assess i) and iii) by examining the sensitivity of the contact solvers with respect to respectively the stopping criterion value $\epsilon_{\text{abs}}$ and the time-step $\Delta t$ (Sec.~\ref{sec:exp_cons}).
Sec.~\ref{sec:exp_perf} evaluates their computational efficiency.
Finally, Sec.\,\ref{sec:exp_mpc} explores how the contact models and their implementations can impact the final robotics applications, in the case of the MPC for quadruped locomotion.

Except where expressly indicated, we use the following values for the solvers' parameters: an absolute convergence criterion $\epsilon_{abs} =10^{-6} $, a maximum number of iteration $ n_{iter} = 10^4$ and a time-step $\Delta t = 1ms$.
For the ADMM algorithm, the proximal parameter $\rho$ is adapted dynamically as previously detailed (Sec.~\ref{sec:implementation_details}).

\subsection{Evaluation of physical correctness} 
\label{sec:exp_phys}

\begin{figure}
    \centering
    \includegraphics[width=.49\linewidth]{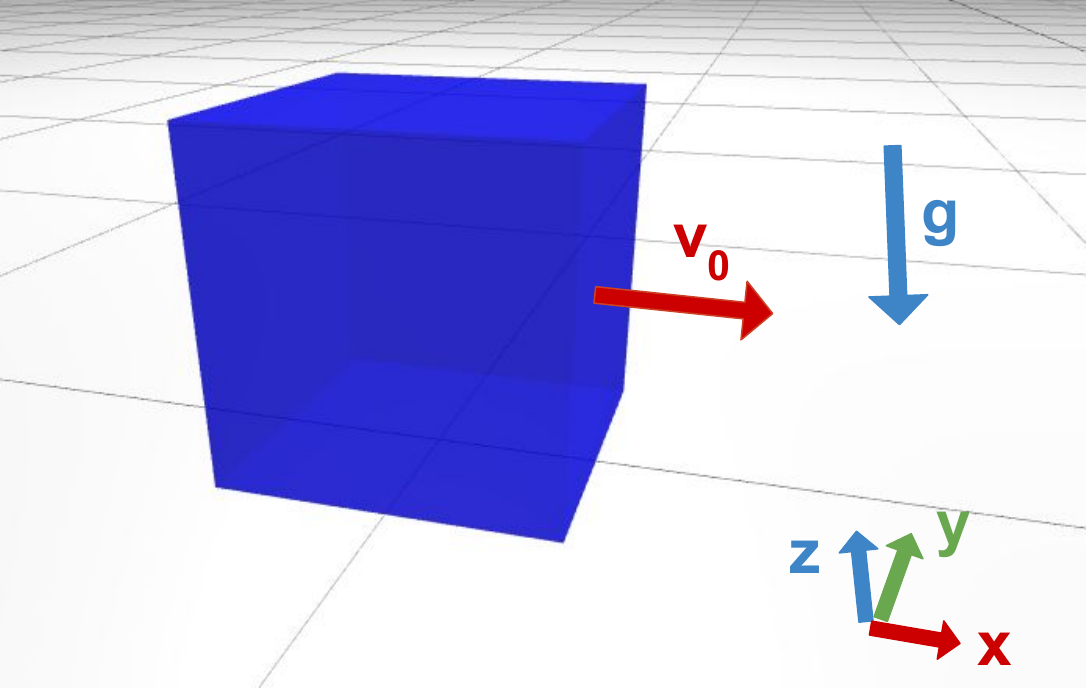}
    \includegraphics[width=.49\linewidth]{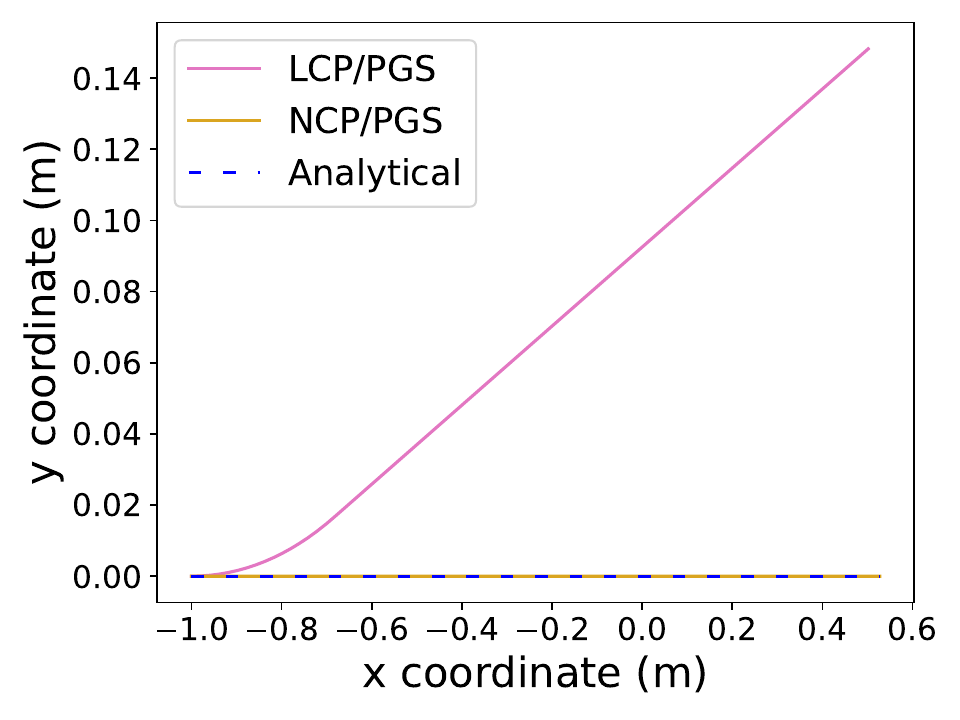}
    \caption{\textbf{Trajectory of a cube sliding on a plane.} The cube is initialized with an initial tangential velocity along the x-axis. 
    \textbf{Right:} The bias of friction forces (Fig.\ref{fig:lcp_bias}) introduces a tangential velocity along the y-axis, which deviates the cube from the expected straight-line trajectory.
    }
    \label{fig:plots_slidingcube_lcp}
    \vspace{-0.0cm}
\end{figure}

\noindent
\textbf{\\LCP relaxation.} 
The linearization of the friction cone loses the isotropy and biases the friction forces towards some specific directions, as shown in Fig.~\ref{fig:lcp_bias}.
This observation has already been raised in the literature~\cite{trinkle1997dynamic,renouf20053d,acary2008numerical,lelidec2021differentiable,howelllecleach2022}.
As expected, the bias on the contact forces significantly impairs the simulation by deviating the trajectory of the simulated system (Fig.~\ref{fig:plots_slidingcube_lcp}).

\noindent
\textbf{\\CCP relaxation.} 
\begin{figure}
    \centering
    \includegraphics[width=.45\linewidth]{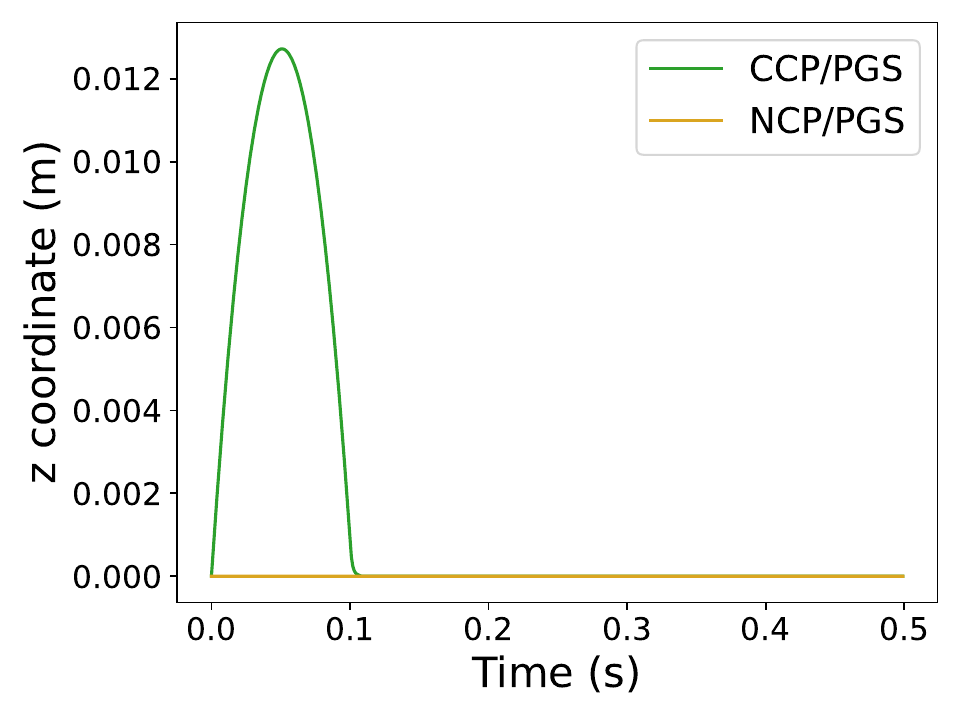}
    \includegraphics[width=.45\linewidth]{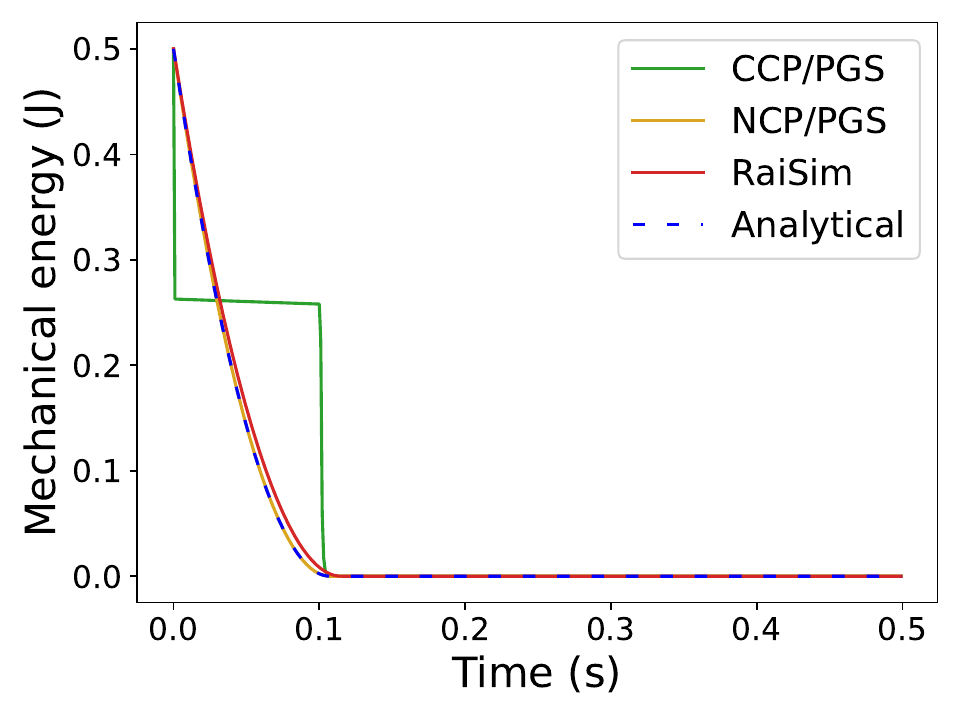}
    \caption{\textbf{A cube is initialized on a plane with a tangential velocity along the x-axis,} similarly to the case studied in Fig.~\ref{fig:plots_slidingcube_lcp}. 
      \textbf{Left:}~The CCP contact model relaxes the Signorini condition, which induces unphysical forces leading to the vertical bouncing of the cube.
     \textbf{Right:}~From the MDP, it is possible to determine the evolution of the energy of the system analytically and compare it to what is computed by the various simulation algorithms.
     The CCP relaxation induces a significant gap with the analytical solution.
     The RaiSim contact model narrows this gap but dissipates less power than expected, as it does not always enforce the MDP.
     The NCP formulation, solved using the PGS solver, is the only formulation that closely matches the expected analytical behavior of the system.
    }
    \label{fig:plots_slidingcube_ccp}
    \vspace{-0.0cm}
\end{figure}
\begin{figure}
    \centering
    \includegraphics[width=.45\linewidth]{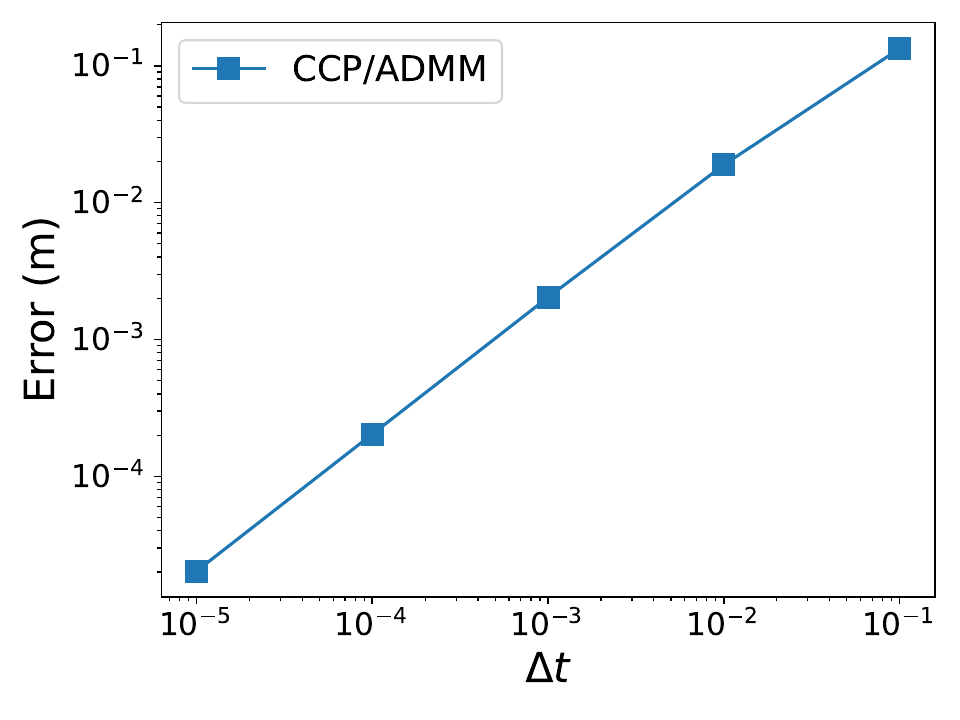}
    \caption{\textbf{Applying a linearly growing force along the x-axis to a cube on a plane.}
    The cube has a mass of 1kg, a side length of 0.2m, a friction coefficient of 0.4 and the external force grows linearly from 0 to 20N over 1s.
    This induces relatively high velocities in a robotic context that are useful for illustrative purposes here.
    }
    \label{fig:plots_slidingcube_ccpncp}
    \vspace{-0.0cm}
\end{figure}
As detailed previously, the CCP contact model relaxes the \textit{Signorini condition}.
As shown in Fig.~\ref{fig:plots_slidingcube_ccp}, this results in non-null normal contact forces and velocities when a contact point is sliding.
As a consequence, the contact points start to bounce, which modifies the trajectory of the system (Fig.~\ref{fig:plots_slidingcube_ccp}, left), which also impacts the overall dissipated energy (Fig.~\ref{fig:plots_slidingcube_ccp}, right).
The model adopted by Raisim aims at correcting this undesired phenomenon by enforcing the Signorini condition but still does not match the analytical solution due to its relaxation of the MDP \eqref{eq:Raisim_kkt} (Fig.~\ref{fig:plots_slidingcube_ccp}, right).

In Figure~\ref{fig:plots_slidingcube_ccpncp}, we simulate a cube dragged on a plane and measure the integral error between the trajectory obtained with the CCP/ADMM solver with various $\Delta t$ and a reference one computed via the NCP/PGS solver with small time step $\Delta t= 10^{-5}s$.
The deviation from the reference trajectory is quantified via the integral consistency error defined as $ \sum_{\tau=0}^T \|  q^\tau - \overline{q}^\tau\| \Delta t$.
As detailed in Sec.~\ref{subsec:ccp}, the error between the trajectories obtained with the CCP and NCP models is proportional to the time step $\Delta t$.

\begin{figure}
    \centering
    \includegraphics[width=.49\linewidth]{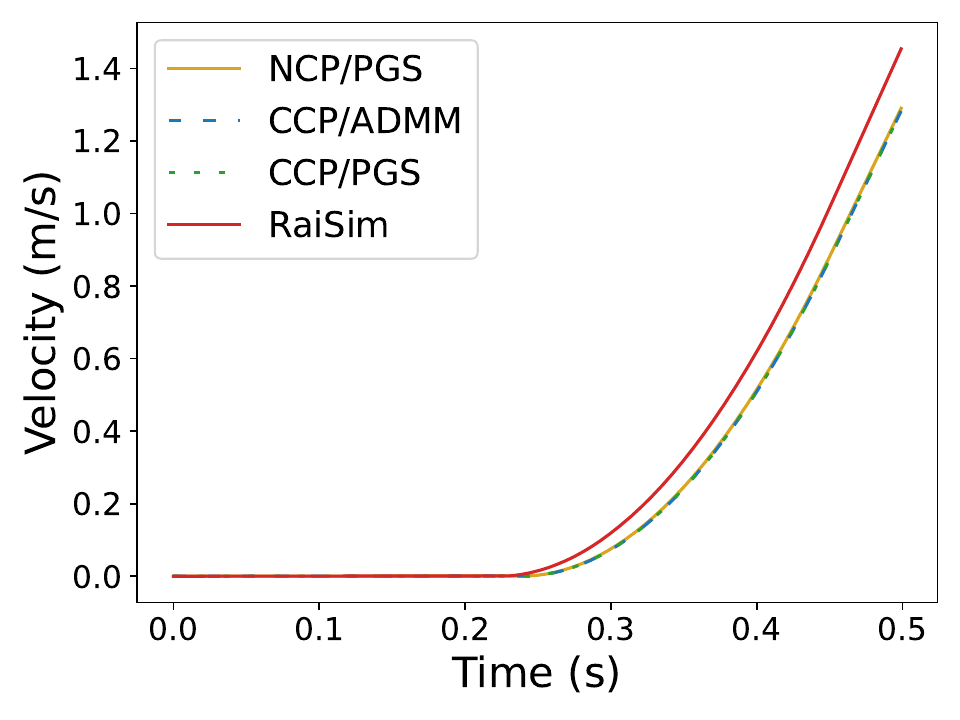}
    \includegraphics[width=.49\linewidth]{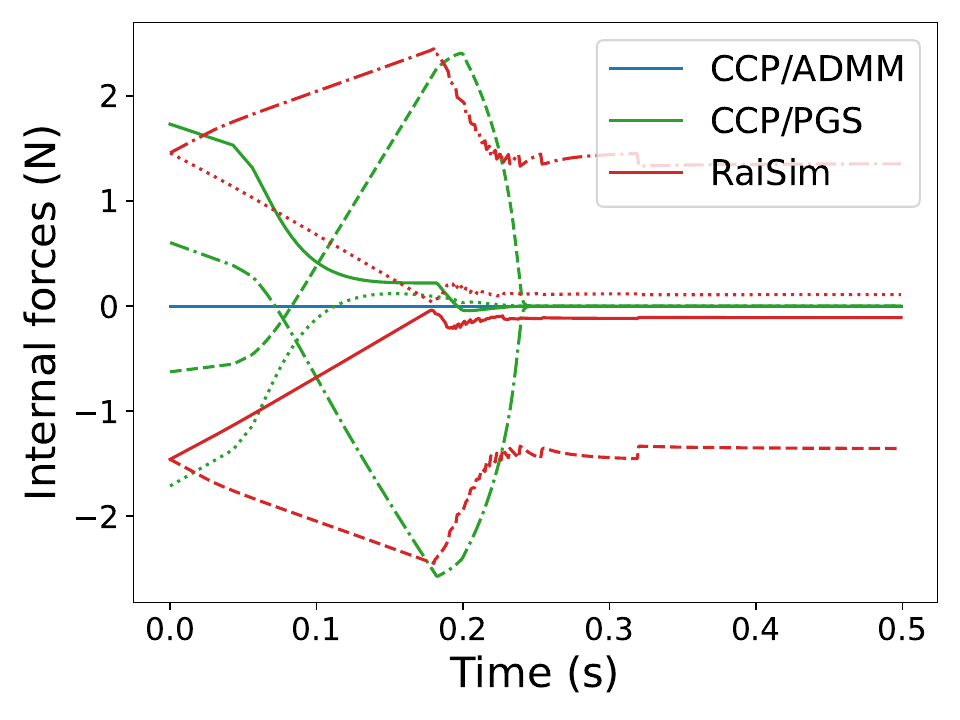}
    \caption{\textbf{A cube is dragged on a plane along the x-axis} similarly to the case studied in Fig.~\ref{fig:plots_slidingcube_ccpncp}.
    \textbf{Left:} The cube is at stiction before it starts sliding after approximately 0.25s.
    The tangential velocity differs depending on the contact model e.g. RaiSim violates the MDP leading to contact points sliding faster than in the case of NCP.
    \textbf{Right:} At stiction, multiple combinations of tangential forces may lead to the same trajectory. 
    There are four curves for each contact model, each curve accounting for the y-component from one of the four contact forces on the cube.
    Gauss-Seidel-like solvers, \textit{e.g.} RaiSim and PGS, exhibit internal forces "stretching" the cube at stiction before the MDP enforces these forces to disappear when the cube starts to slide.
    RaiSim relaxes the maximum dissipation principle so the friction forces are not opposed to the movements, and internal forces persist when the cube is sliding.
    Eventually, ADMM avoids injecting jamming internal forces even at stiction.
    }
    \label{fig:internal_forces}
    \vspace{-0.0cm}
\end{figure}

\noindent
\textbf{\\Underdetermined contact problems.} 
Underdetermination occurs when infinite combinations of contact forces lead to the same trajectory. 
These artifacts happen on the normal and tangential components of contact forces, as depicted in Fig.~\ref{fig:internal_forces}.
As shown in Fig.~\ref{fig:internal_forces}, the solution found depends on the numerical scheme.
We observe that the per-contact approaches (Alg.~\ref{alg:ccppgs},\ref{alg:raisim} and \ref{alg:ncppgs}) exhibit jamming internal forces at stiction, values which are not controlled by the algorithms.
On the opposite, the algorithms working directly on the global contact problem with a proximal regularization (Alg.~\ref{alg:ccpadmm} and \ref{alg:stag_proj}) seem to avoid injecting such artifacts in the contact forces (Fig.~,\ref{fig:internal_forces}).
As future work, it would be interesting to investigate the theory behind the latter conjecture.

This phenomenon may seem innocuous as forward dynamics are not affected.
However, it makes the inverse dynamics ill-posed, as there is no way to predict such numerical artifacts.
Additionally, in the context of differentiable physics, we believe these spurious contact forces may catastrophically impact the computation of derivatives, but we also leave this study as future work.
Finally, it is worth mentioning that such under-determined cases are ubiquitous in robotics (e.g., legged robots making redundant contact with their environments).

\begin{figure}[h]
    \centering
    \includegraphics[width=.49\linewidth]{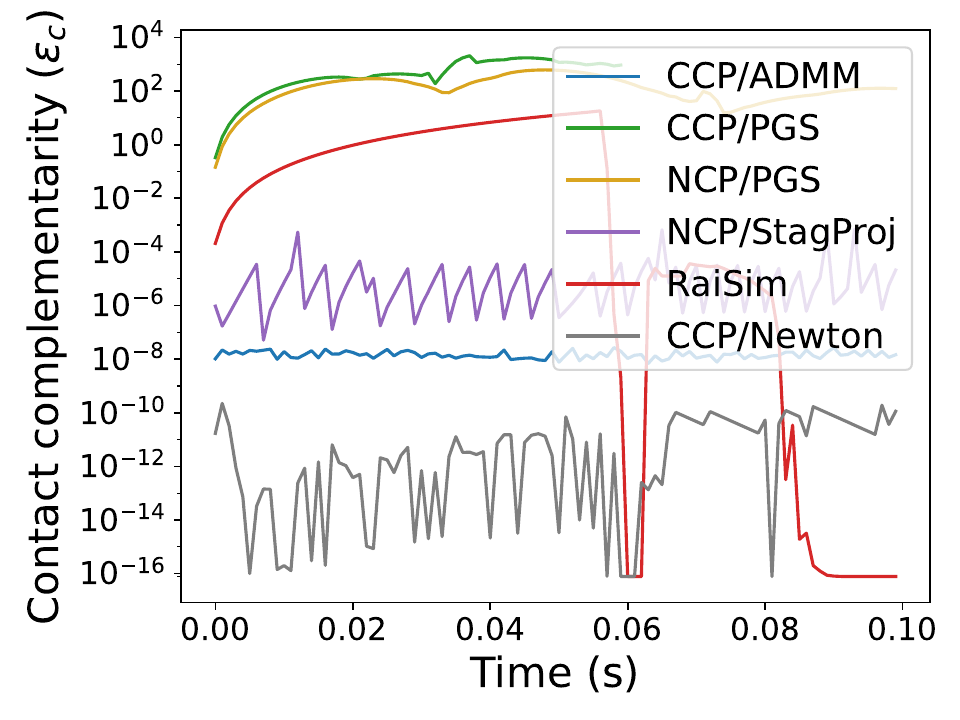}
    \includegraphics[width=.49\linewidth]{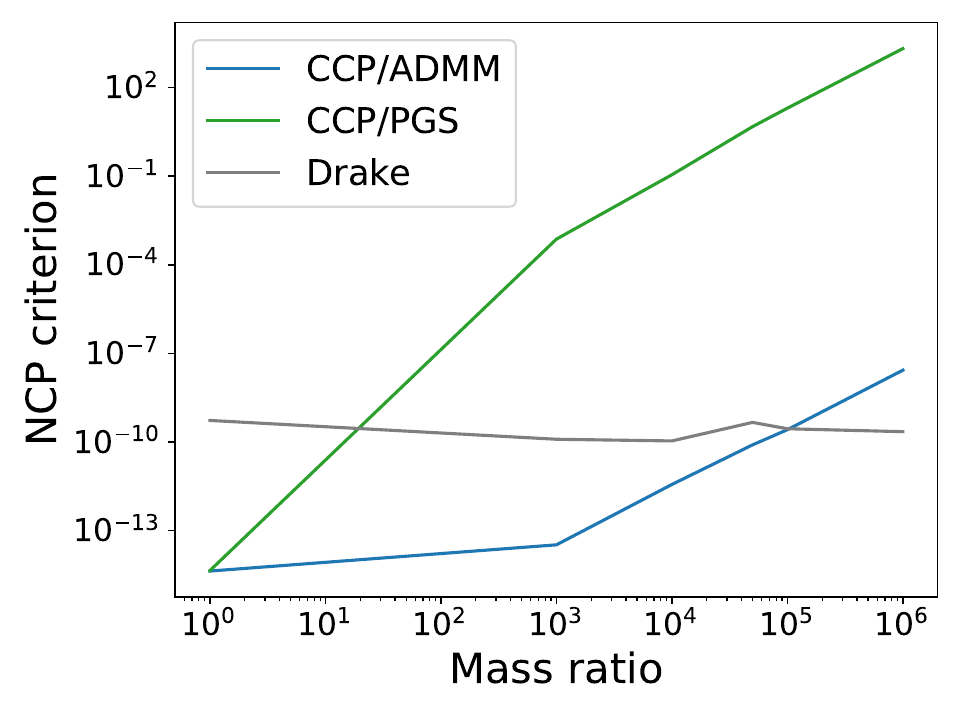}
    \caption{\textbf{Simulation of ill-conditioned systems.}
    \textbf{Left:} Stacking a heavy cube ($10^3$kg) on a light one ($10^{-3}$kg) makes the problem ill-conditioned and, therefore, not solvable via per-contact algorithms (CCP/PGS, NCP/PGS and RaiSim) which results in the violation of the contact complementarity criterion \eqref{eq:NCP}.
     By contrast, the ADMM, staggered projections, and Newton approaches appear to be robust in this case.
    \textbf{Right:} The accuracy of the simulators improves when the ratio between the masses of the two cubes gets close to one.
    The ADMM and Newton algorithms are less affected by this ratio than PGS.}
    \label{fig:plots_2cubes}
    \vspace{-0.0cm}
\end{figure}

\noindent
\textbf{\\Robustness to ill-conditioned contact problems.} More generally, the contact problem becomes challenging when the ratio between the biggest and the smallest eigenvalue of the Delassus matrix grows.
The experiment of Fig.~\ref{fig:plots_2cubes} exhibits the convergence issues of per-contact approaches when simulating systems with a strong coupling between the different contact points, which causes large off-diagonal terms on the matrix $G$.
In this situation, the latter approaches hit the maximum number of iterations before convergence, leading to unrealistic trajectories.
Such a behavior can be expected as supposing the matrix to be diagonally dominant is a classical hypothesis ensuring the convergence of Gauss-Seidel methods.
On the contrary, the proximal algorithms account for off-diagonal terms of $G$, and only rely on a regularized inverse of $G$ (Alg.~\ref{alg:ccpadmm}, line~\ref{line:reg_inverse}), and thus robustly converge towards an optimal solution.

\begin{figure*}[h]
    \centering
    \includegraphics[width=.25\linewidth]{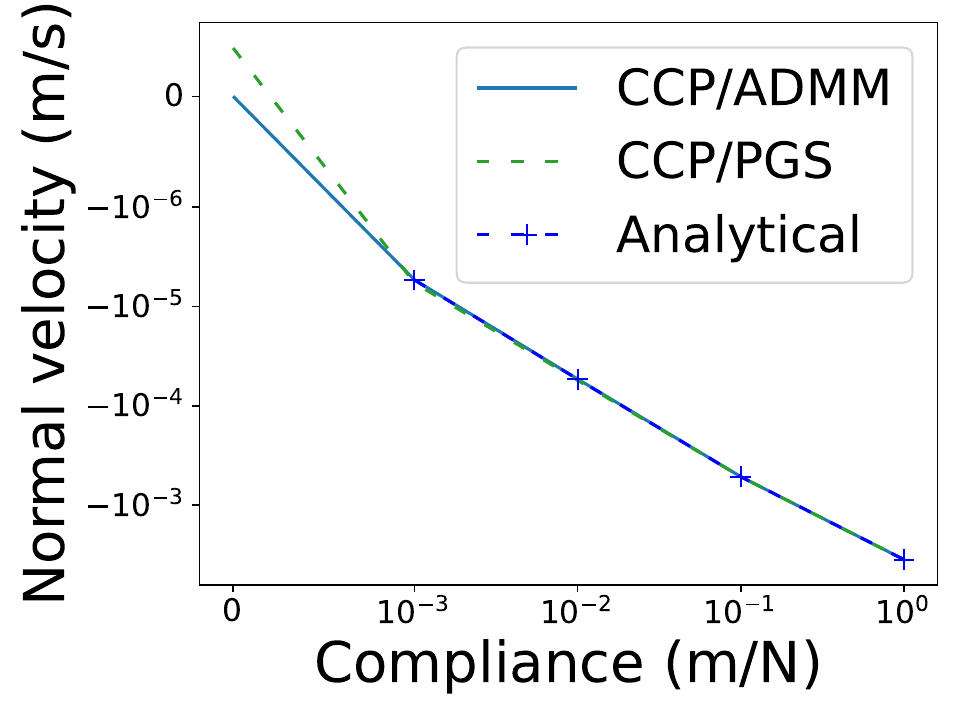}
    \includegraphics[width=.25\linewidth]{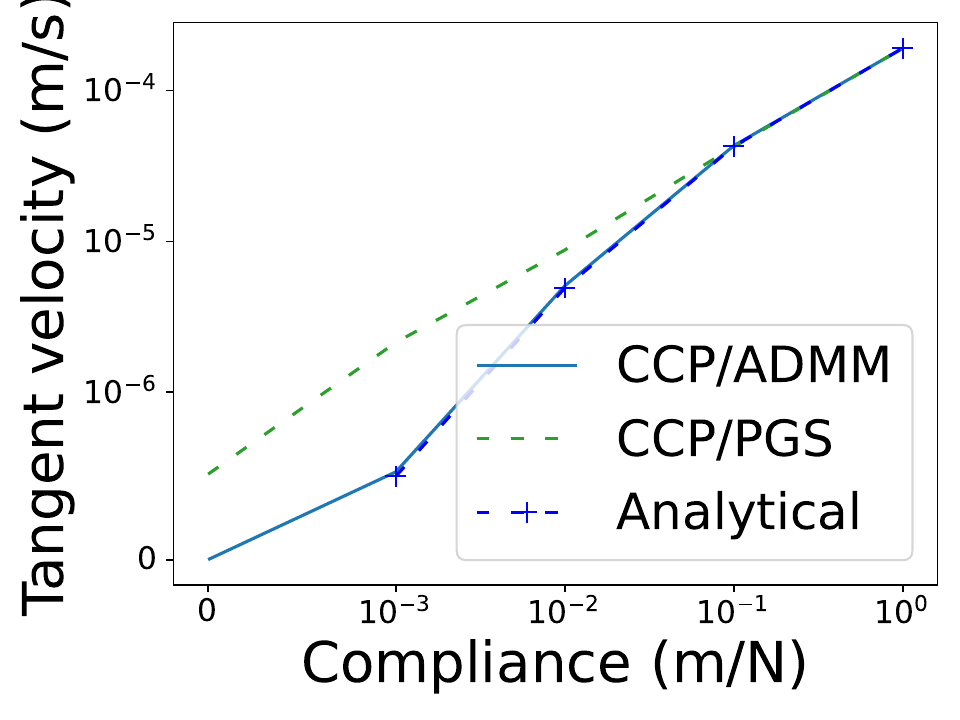}
    \includegraphics[width=.25\linewidth]{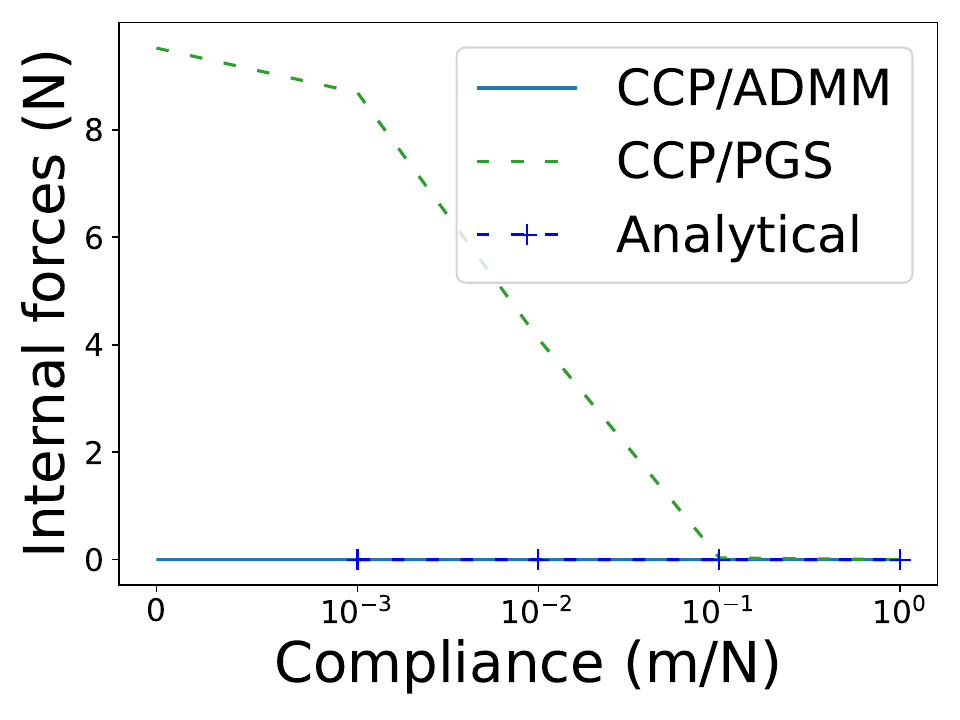}
    \caption{\textbf{Simulation of Solo-12 with varying compliance for the contacts with the floor.}
    The robot is in a standing position and perturbed with an external horizontal force.
     \textbf{Left:} Adding a compliance to the contact model relaxes the Signorini condition.
     \textbf{Center:} This compliance also relaxes the Coulomb's law of friction.
     \textbf{Right:} Compliance also regularizes the problem, removing the jamming internal forces in the under-determined cases.}
    \label{fig:compliant_contacts}
    \vspace{-0.0cm}
\end{figure*}

\noindent
\textbf{\\Effects of compliance.}
As demonstrated by Fig.~\ref{fig:compliant_contacts}, the normal forces vary linearly with the compliance parameter $R$.
Moreover, adding compliance to the tangential components induces the vanishing of dry friction, resulting in tangential oscillations instead of a null velocity.
These compliant effects regularize the infinitely steep graphs due to the Signorini condition and Coulomb's law and replace them with locally linear mapping, which also eases the numerics.
Therefore, the compliance added in MuJoCo has no physical purpose and should be considered a numerical trick designed to circumvent the issues due to hyper-staticity or ill-conditioning at the cost of impairing the simulation.

\subsection{Self-consistency of the solvers}\label{sec:exp_cons}
The accuracy of simulators can be affected by the numerical resolution induced by two "hyper-parameters": the value of the stopping criterion for the contact 
solver desired accuracy~($\epsilon_{abs})$ and the time-step value ($\Delta t$).
We measure their effect on the simulation quality when varying them independently.
A simulator is said to be self-consistent when this deviation remains limited.

\begin{figure}[h]
    \centering
    \includegraphics[width=.6\linewidth]{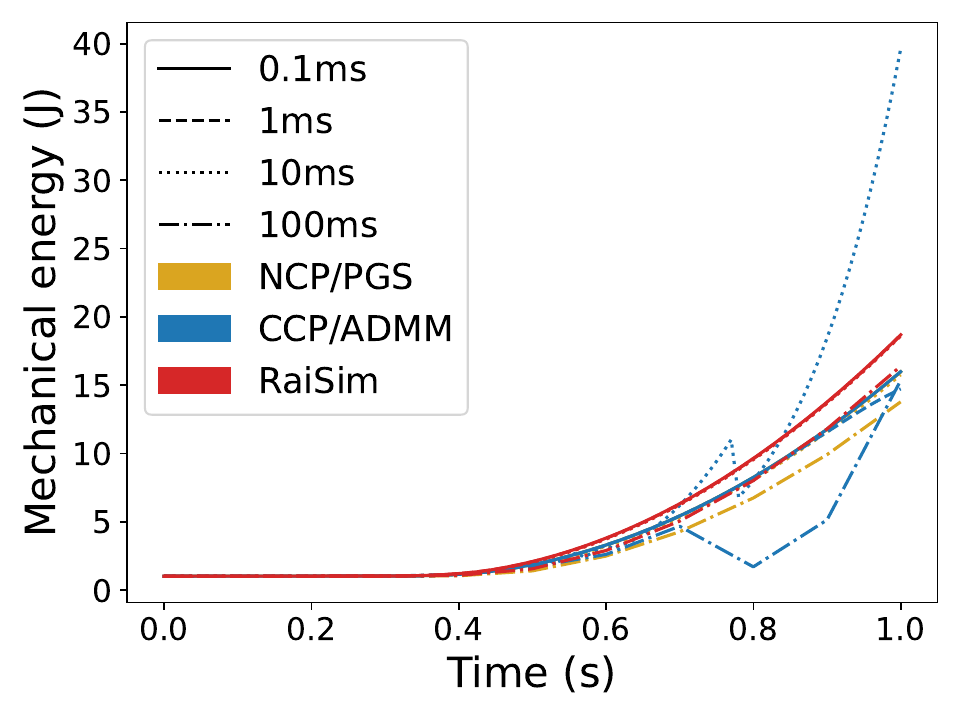}
    \includegraphics[width=.6\linewidth]{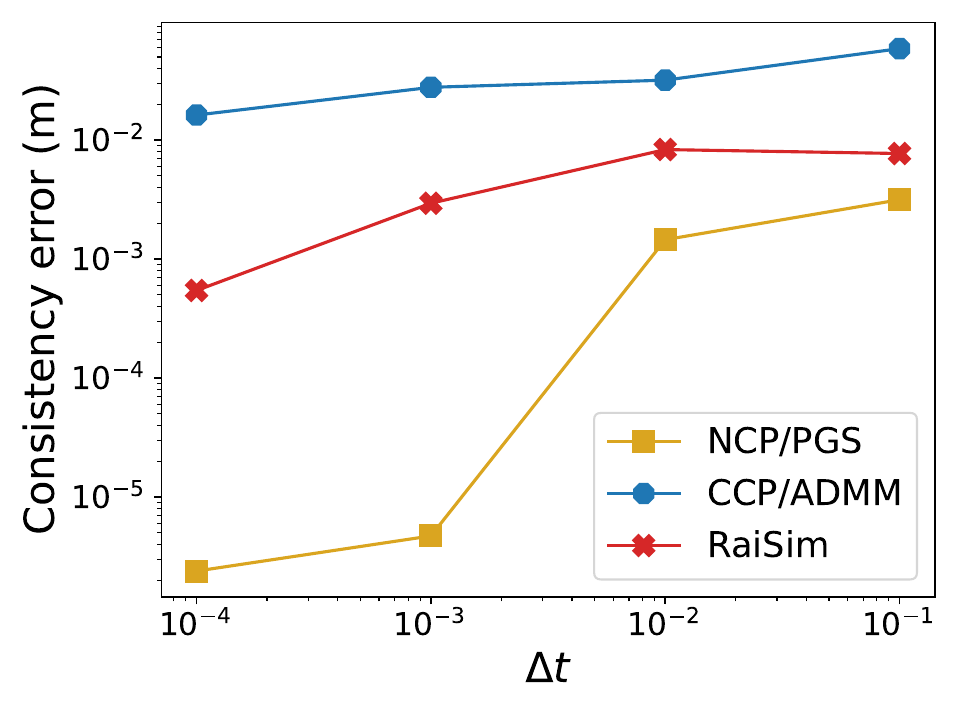}
    \caption{\textbf{Self-consistency w.r.t. time-stepping when simulating a cube dragged on a plane by a growing tangential force.}
    The CCP contact model appears to be more sensitive to the time step $\Delta t$.
    This sensitivity can also be observed through the evolution of mechanical energy.}
    \label{fig:self_consistency_dt}
    \vspace{-0.0cm}
\end{figure}

\noindent
Time-stepping simulators are sensitive to the choice of the time-step $\Delta t$.
Here, we intend to assess the self-consistency of the various contact solvers by examining their deviation when $\Delta t$ grows.
Because time discretization also affects the collision detection process, our study is done on the trajectory of a cube dragged on a plane by a growing tangential force and whose contact points should remain constant (as done in Fig.~\ref{fig:plots_slidingcube_ccpncp}).
This scenario also allows to asses both sticking and sliding modes.
For each simulator, a trajectory $\overline{q}$ obtained by simulating the system with a small time-step ($\Delta t =10^{-2}ms$) serves as a reference to compute the state consistency error along the trajectories simulated with larger time-steps (Fig.~\ref{fig:self_consistency_dt}).

Looking at Fig.~\ref{fig:self_consistency_dt}, we observe that the CCP contact model is more sensitive with respect to the time step in the considered scenario.
Indeed, because CCP relaxes the \textit{Signorini condition}, the cube slides at a height proportional to $\Delta t$.
Similarly, as shown by Fig.~\ref{fig:self_consistency_dt}, the energy evolution of the system simulated via NCP and RaiSim models is only a little modified when increasing $\Delta t$ while CCP leads to a nonphysical and inconsistent behavior.

\subsection{Performance benchmarks} 
\label{sec:exp_perf}

\begin{figure}[h]
    \centering
    \includegraphics[width=.45\linewidth]{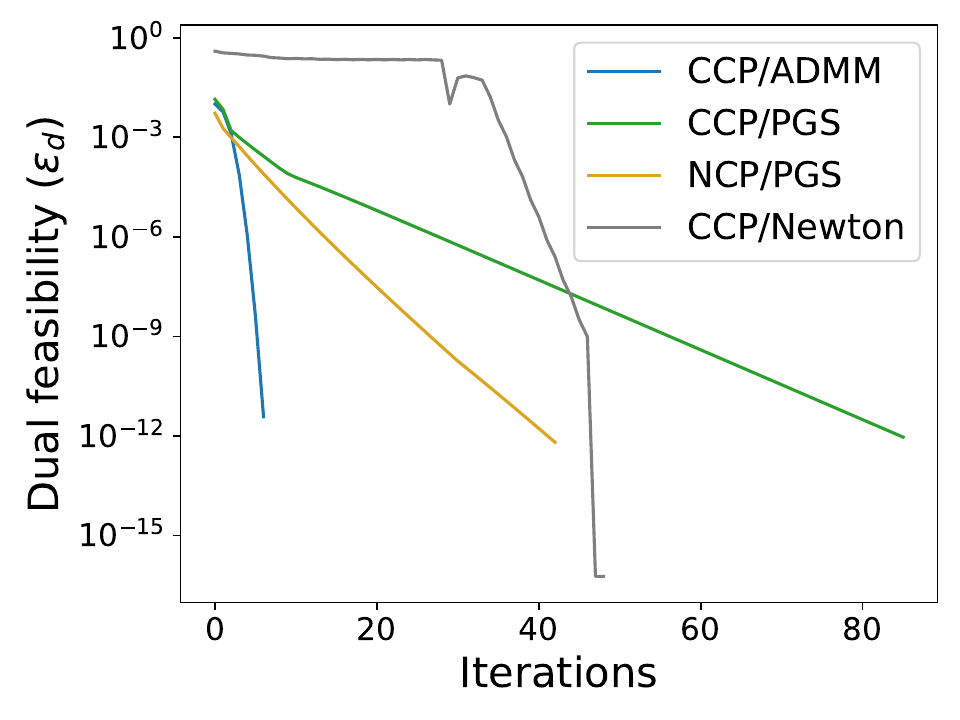}
    \includegraphics[width=.45\linewidth]{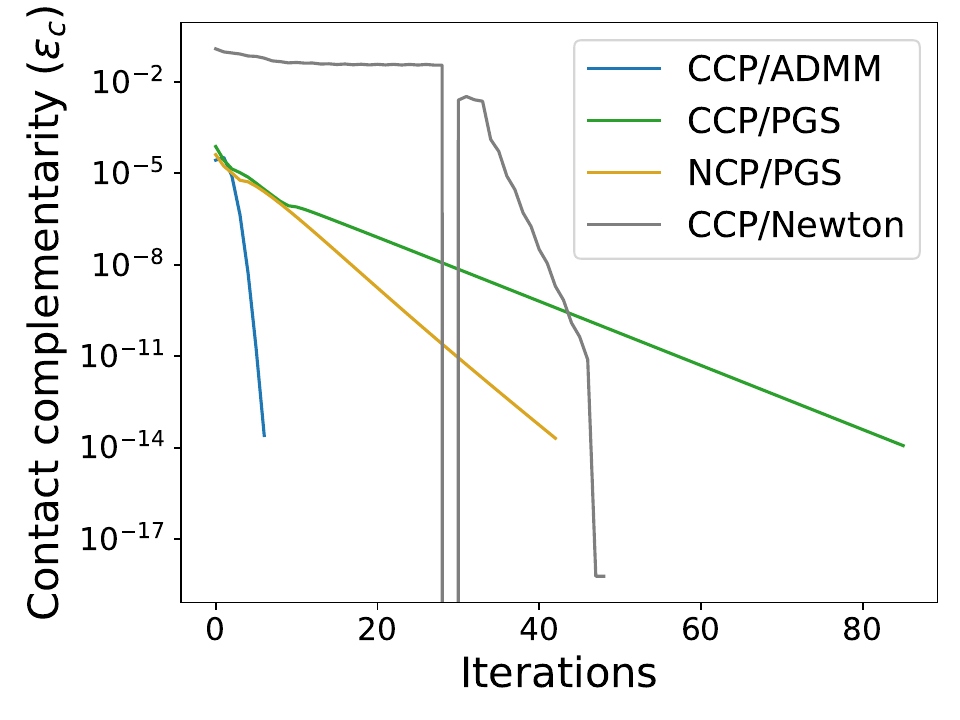}
    \includegraphics[width=.45\linewidth]{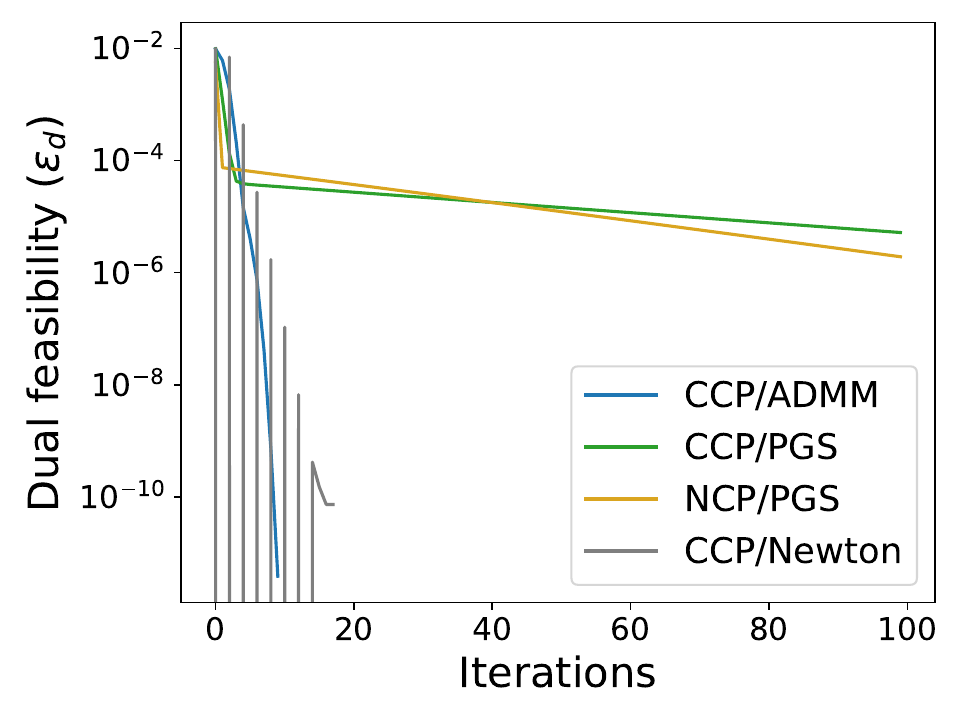}
    \includegraphics[width=.45\linewidth]{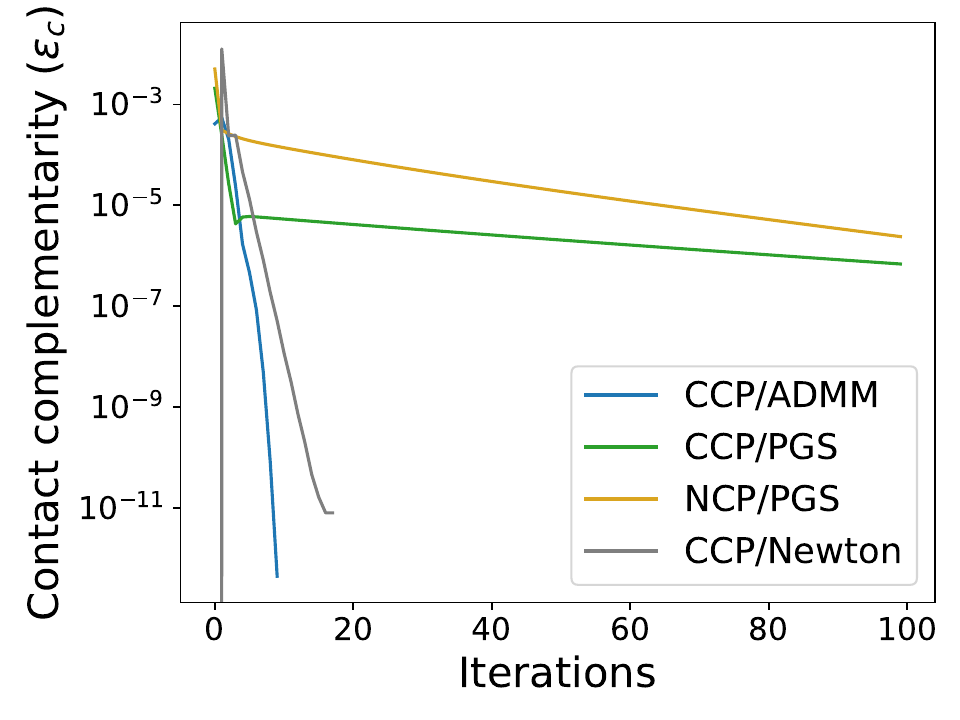}
    \caption{\textbf{Convergence for contact problems on Solo (Top) and Talos (Bottom) robots.}
    The considered contact problems are extracted from one time step of the full trajectory.
    PGS is fast to reach a mild accuracy before saturating, while ADMM and second-order algorithms can get to higher precision.}
    \label{fig:iterations}
    \vspace{-0.0cm}
\end{figure}

\begin{figure}[h]
    \centering
    \includegraphics[width=.45\linewidth]{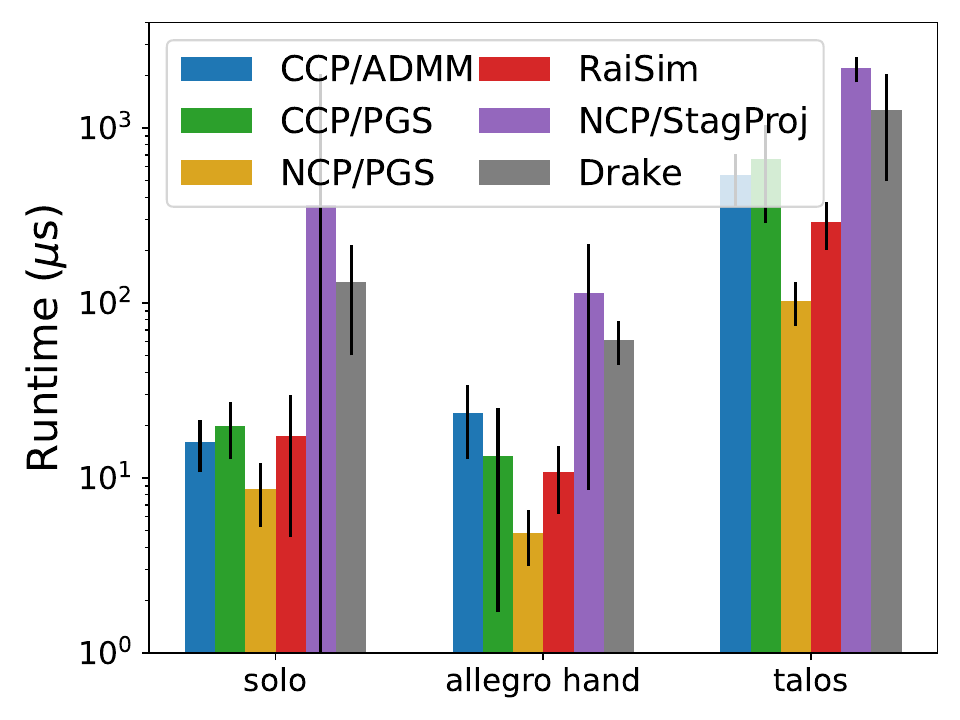}
    \includegraphics[width=.45\linewidth]{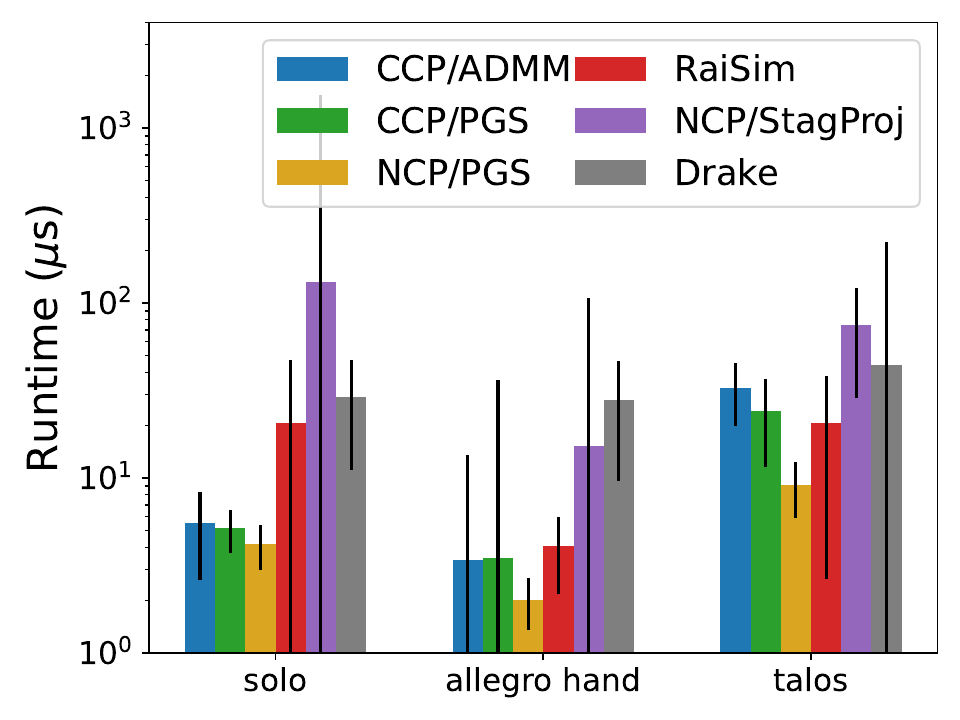}
    \caption{\textbf{Computational timings measured along a trajectory for three robotic systems} (\textit{c.f.} Fig.~\ref{fig:robot_exp}).
    The represented timings are obtained by averaging on the entire trajectory.
    The contact solvers are tested in both cold-start (\textbf{Left}) and warm-start (\textbf{Right}) modes.
    We simulate the same trajectories to evaluate the benefit of warm-starting, but we use the solution of the previous time step as an initial guess. 
    This leads to significant improvements in the computational timings.}
    \label{fig:timings}
    \vspace{-0.0cm}
\end{figure}

As evoked earlier, in addition to being physically accurate, it is also essential for a simulator to be fast, which, in general, constitutes two adversarial requirements.
To evaluate the computational footprint of the various solvers, we measure both the number of iterations and the time taken to reach a fixed accuracy on dynamic trajectories involving robotics systems (Fig.~\ref{fig:robot_exp}).
This is done on three different robotics scenarios: the quadruped Solo (12-dof) and the humanoid Talos (32-dof) are in a standing position and perturbed by applying an external force (of respectively 10N and 80N) at their center of mass, while a ball is dropped in the Allegro hand (16-dof), so the trajectories are not static. 

Looking at the number of iterations required to converge (Fig.~\ref{fig:iterations}), PGS approaches appear to be reasonably fast to reach mild accuracy ($\epsilon_{\text{abs}}= 10^{-5}$) while they eventually saturate before reaching high precision in complex scenarii (Fig.~\ref{fig:iterations}, bottom).
We show later this can be insufficient for challenging tasks (Fig.~\ref{fig:mpc_bumpy_terrain}).
On the other hand, ADMM, Newton and Staggered Projections algorithms can find high-accuracy solutions using only a few, but more costly, iterations.
As mentioned earlier, our implementation of Drake's solver uses a backtracking line search with an Armijo condition while the original algorithm \cite{castro2022unconstrained} uses a more advanced routine.
This difference could lead to degraded performances here which should be kept in mind when interpreting the results.

The latter analysis does not account for the per-iteration computational cost, so we report a study on final timings in Fig.~\ref{fig:timings}.
As they work with various contact model hypotheses and thus have different convergence criteria, it is challenging to make a fair comparison between the solvers.
Therefore, we choose to run the solvers in the setup they are usually used in practice: the solver is stopped whenever it reaches an absolute convergence threshold ($\epsilon_\text{abs} = 10^{-6}$) or otherwise, it is early-stopped if a maximum number of iteration ($n_\text{iter} = 10^{4}$) is reached or stalling is detected via relative convergence criterion ($\epsilon_\text{abs} = 10^{-8}$).
For this reason, Fig.~\ref{fig:timings} is only informative about the computational cost but not about the accuracy of solvers.
When the contact solvers are cold started, we observe that the second-order optimization techniques \cite{todorov2012mujoco,castro2022unconstrained} are less efficient than the PGS solvers and their cheap per-contact iterations (Fig.~\ref{fig:timings}, left).
The advanced first-order algorithms like ADMM (Alg.~\ref{alg:ccpadmm}) working on the dual CCP problem \eqref{eq:CCP_QCQP} stands in-between as they leverage the very efficient Featherstone algebra \cite{carpentier2021proximal} for the computation of the Cholesky factorization of $G$ (Sec.~\ref{sec:implementation_details}).
However, leveraging the solution from the previous time step to warm-start the solvers --- a common strategy in practice --- allows for significantly reducing this gap (Fig.~\ref{fig:timings}, right).
Therefore, regarding the study of Sec.~\ref{sec:exp_phys}, a trade-off appears for algorithms like ADMM, which treat all the contact points globally. 
In practice, they might be slower than their PGS counterpart while they benefit from better behaviors on ill-conditioned problems.

\subsection{MPC for quadruped locomotion} \label{sec:exp_mpc}

The previous examples already illustrate the differences among the various simulators in terms of both physical accuracy and computational efficiency.
However, such scenarios may not represent the richness of contacts in practical robotics situations.
For this purpose, we use the implementation of MPC on the Solo-12 system introduced in \cite{leziart2021qrw} to generate locomotion trajectories on flat and bumpy terrains.
These experiments are designed to involve a wide variety of contacts (\textit{i.e.}, sticking, sliding, and taking-off) and see how the simulation choices impact the final task (\textit{i.e.}, horizontal translation of the robot).

\begin{figure}[h]
    \centering
    \includegraphics[width=.45\linewidth]{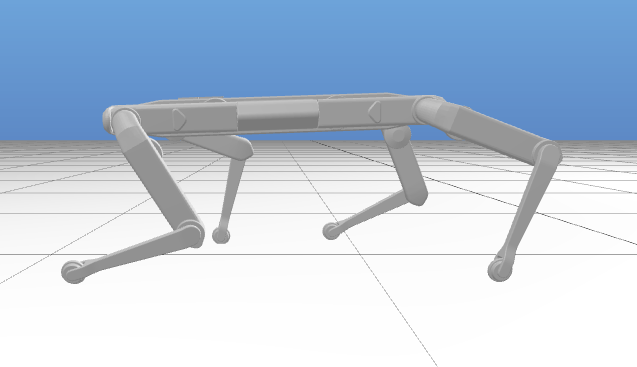}
    \includegraphics[width=.45\linewidth]{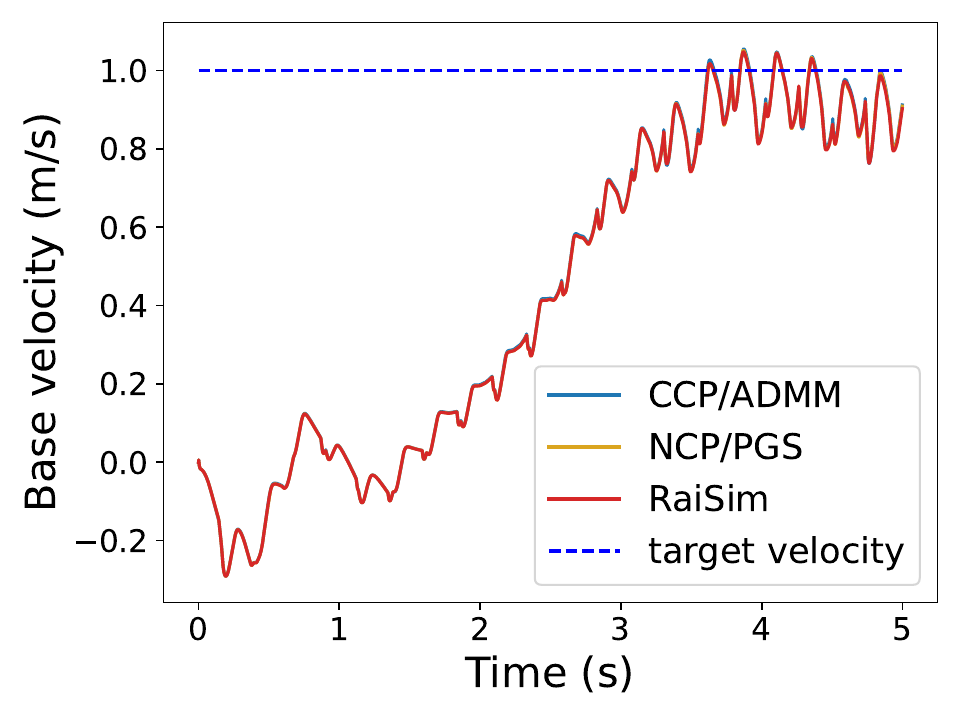}
    \includegraphics[width=.45\linewidth]{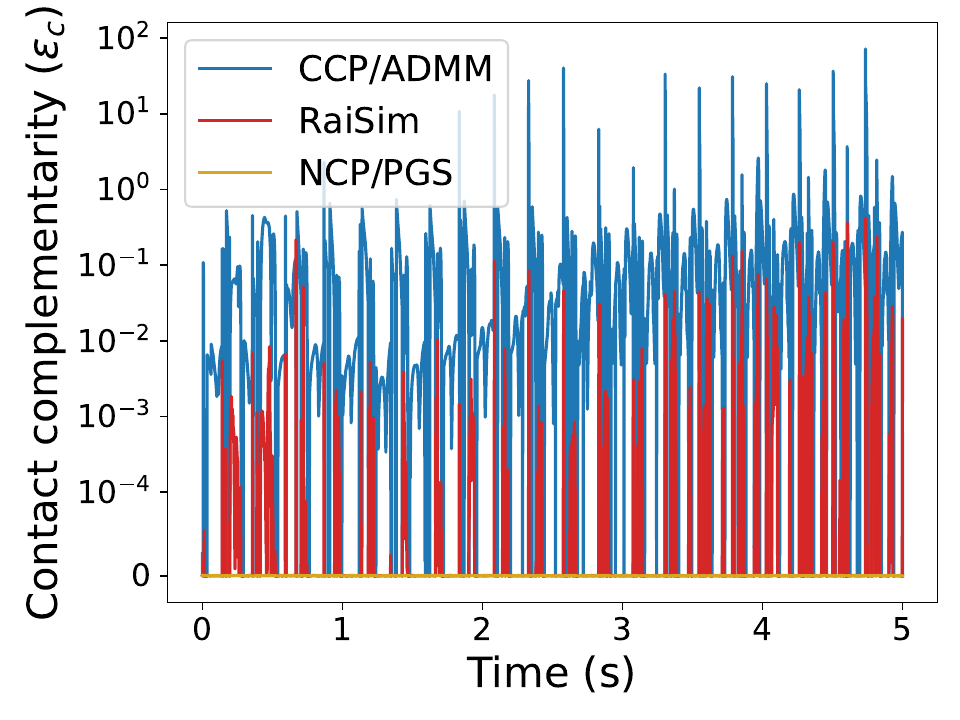}
    \includegraphics[width=.45\linewidth]{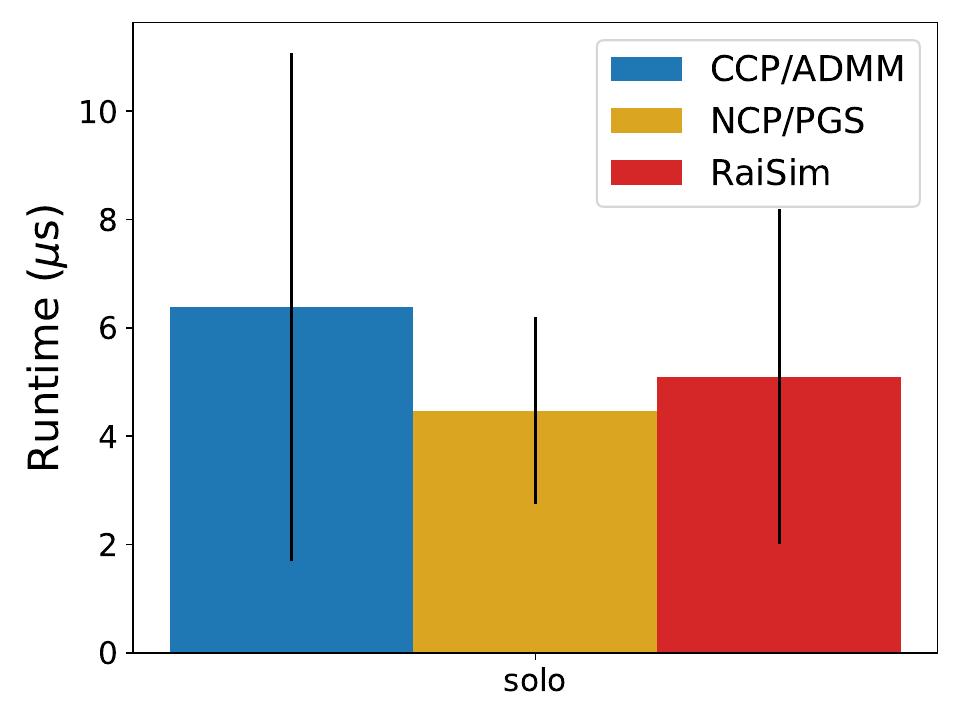}
    \caption{\textbf{MPC for locomotion on a flat terrain} (\textbf{Top left}).
    The target horizontal translation velocity of the base is similarly reached by the controller with the different simulators (\textbf{Top right}).
    However, they do not equally respect the contact complementarity criterion \eqref{eq:NCP} (\textbf{Bottom left}).
    Per-contact approaches, \textit{e.g.} PGS and RaiSim, are more efficient (\textbf{Bottom right}).}
    \label{fig:mpc_flat_terrain}
    \vspace{-0.0cm}
\end{figure}

For a flat and barely slippery ($\mu=0.9$) ground, we observe that the choice of simulator hardly affects the base velocity tracked by the MPC controller (Fig.~\ref{fig:mpc_flat_terrain}, top right). 
In this case, the contacts are mainly sticking, leading to low violation of the NCP criterion \eqref{eq:NCP} (Fig.~\ref{fig:mpc_flat_terrain}, bottom left).

\begin{figure}[h]
    \centering
    \includegraphics[width=.45\linewidth]{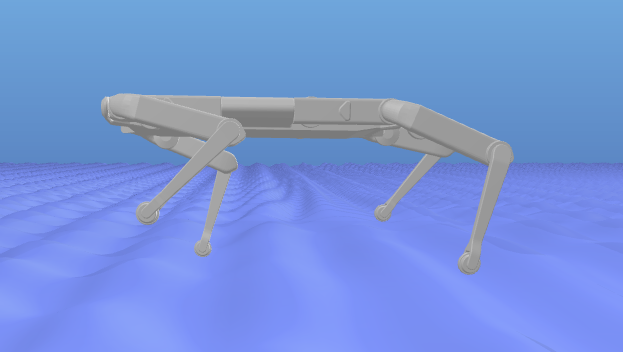}
    \includegraphics[width=.45\linewidth]{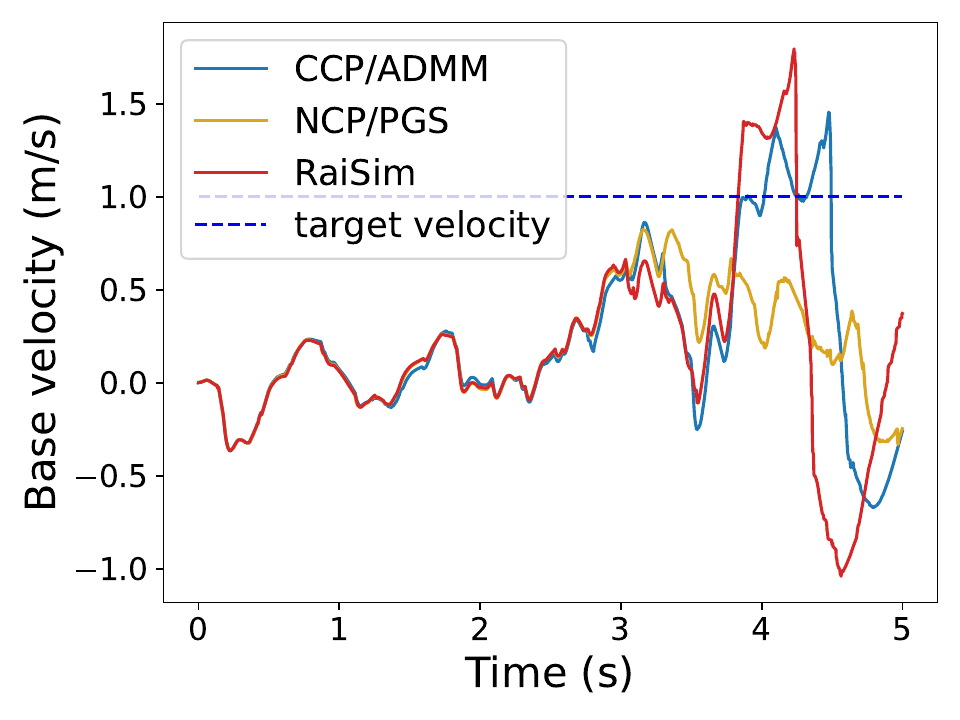}
    \includegraphics[width=.45\linewidth]{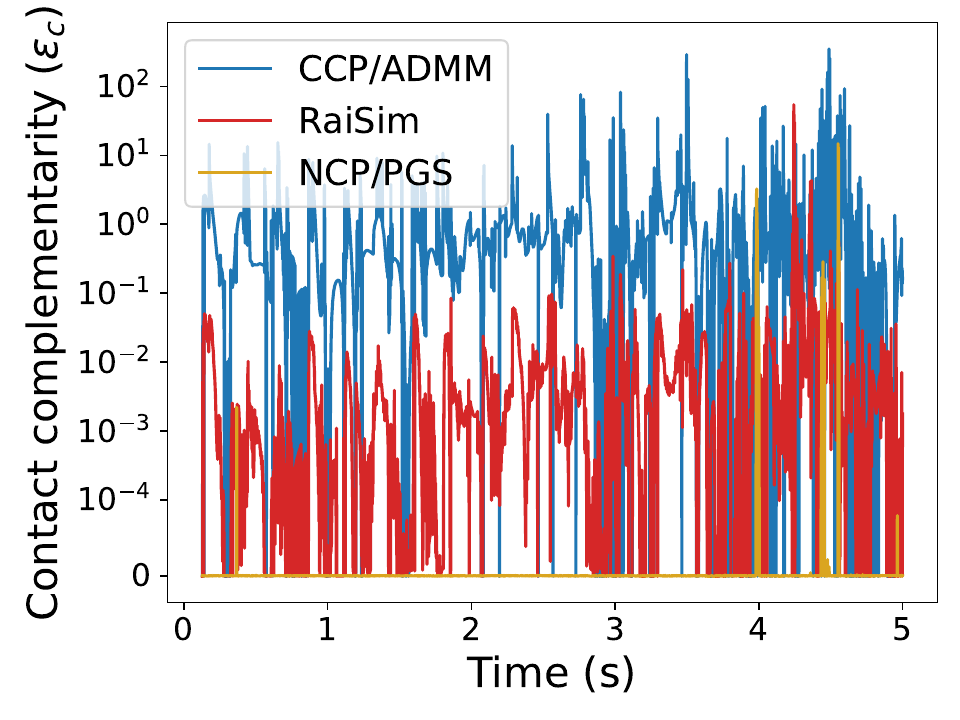}
    \includegraphics[width=.45\linewidth]{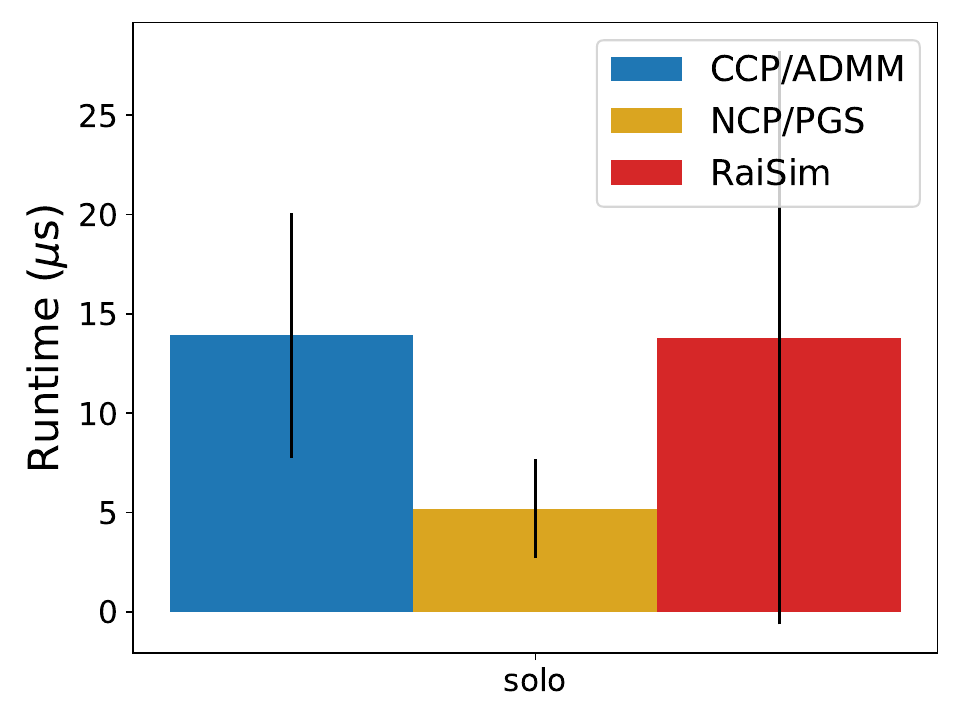}
    \caption{\textbf{MPC for locomotion on a bumpy terrain} (\textbf{Top left}).
    The tracked velocity of the base quickly differs depending on the used simulator (\textbf{Top right}).
    Slippery contact points violate the contact complementarity criterion \eqref{eq:NCP} for the RaiSim and CCP contact modelings (\textbf{Bottom left}).
    The complexity of contacts also hampers the solvers and reduces the gap between per-contact and ADMM approaches (\textbf{Bottom right}).}
    \label{fig:mpc_bumpy_terrain}
    \vspace{-0.0cm}
\end{figure}

When the terrain is bumpy (roughness of $10^{-1}m$) and slippery ($\mu= 0.3$), the locomotion velocity generated from the RaiSim and CCP models significantly deviates from the NCP one (Fig.~\ref{fig:mpc_bumpy_terrain}, top right).
This can be expected, in light of our previous study, as both the RaiSim and CCP contact models make physical approximations when contact points are sliding (Fig.~\ref{fig:mpc_bumpy_terrain}, bottom left).
However, we occasionally observe that NCP/PGS also violates the NCP criterion \eqref{eq:NCP} (Fig.~\ref{fig:mpc_bumpy_terrain}, bottom left) but for a different reason: PGS was not able to converge before the maximum number of iterations was reached.
Therefore, Gauss-Seidel-like approaches appear to be sufficient for mild conditions (Fig.~\ref{fig:mpc_flat_terrain}) but are not robust enough to ensure convergence of the simulation when the locomotion tasks become more challenging (Fig.~\ref{fig:mpc_bumpy_terrain}).
This also causes increased computations from the solvers, particularly for RaiSim (Fig.~\ref{fig:mpc_bumpy_terrain}, bottom right).
These observations indicate that the combination of low-level choices on both the contact model and solver may induce significant differences in the high-level behaviors of locomotion controllers on complex terrains.

\section{Discussion and conclusion}

NCP is known to be complex to solve and thus is often relaxed to find approximate solutions.
In this article, we report a deeper study on how the various rigid contact models commonly employed in robotics and their associated solvers can impact the resulting simulation.
We have notably established and experimentally highlighted that these choices may induce unphysical artifacts, thus widening the reality gap, leading to unrealistic behaviors when the simulator is later used for practical robotics applications.
Our experiments show that there is no fully satisfactory approach at the moment, as all existing solutions compromise either accuracy, robustness, or efficiency.
This indicates that there may still be room for improvements in contact simulation.
It is also worth mentionning that, for robotics, simulation samples of lesser but controlled accuracy are already valuable for many applications, \textit{e.g.} RL and MPC, while a failed simulation represent a waste of ressources.
This paper showcases that situations prone to failure of simulation are not only corner cases but can become quite common when adressing challenging tasks such as locomotion.
This justifies the emphasis put by modern simulators \cite{todorov2012mujoco,macklin2019non,castro2022unconstrained} on robustness when modeling contacts and implementing the associated solvers.

Beyond contact simulation, differentiable physics constitutes an emergent and closely related topic.
However, the impact of forward simulation artifacts on gradient computation remains unexplored.
In particular, some of the relaxations at work, \textit{e.g.} the artificial compliance added in MuJoCo, result in crucial differences in gradients, which then affect downstream applications like trajectory optimization \cite{suh2022bundled,lelidec2022leveraging}.
We leave the study of the various existing differentiable simulators \cite{de2018end, lelidec2021differentiable, werling2021fast, heiden2021neuralsim, howelllecleach2022} through this lens as future work.

For all these reasons, we believe it would be highly beneficial for the robotics community to take up such low-level topics around simulation, as they could lead to substantial progress in the field.
The work of \cite{Horak2019OnTS} is an inspiring first step in this direction. 
With this article, we intend to go further by also providing open-source implementations and benchmarks to the community.

\section*{Acknowledgments}
We warmly thank Jemin Hwangbo for providing useful details on the algorithm of the RaiSim simulator, Stéphane Caron and Nicolas Mansard for helpful discussions.
This work was supported in part by L'Agence d'Innovation Défense, the French government under the management of Agence Nationale de la Recherche through the project INEXACT (ANR-22-CE33-0007-01) and as part of the "Investissements d'avenir" program, reference ANR-19-P3IA-0001 (PRAIRIE 3IA Institute), by the  European Union through the AGIMUS project (GA no.101070165) and the Louis Vuitton ENS Chair on Artificial Intelligence.
Views and opinions expressed are those of the author(s) only and do not necessarily reflect those of the European Union or the European Commission. Neither the European Union nor the European Commission can be held responsible for them.

\balance
\bibliographystyle{ieeetr}
\bibliography{bibliography}








\end{document}